\documentclass[letterpaper, 10 pt, conference, onecolumn]{ieeeconf}

\IEEEoverridecommandlockouts                              

\usepackage{acronym}
\usepackage{comment}
\usepackage{siunitx}
\usepackage{tabu}

\usepackage{ifthen}
\usepackage{algorithm}
\usepackage{amsmath, amsfonts, amssymb} 
\usepackage{graphicx, import} 
\usepackage{bm}
\usepackage{wrapfig}
\usepackage[font=small]{caption}
\usepackage{placeins}
\usepackage{hyperref}
\usepackage[normalem]{ulem}
\usepackage{multirow}
\usepackage{booktabs}

\newcommand{\wcnn}{\emph{WindSeer}}
\newcommand{\Perdigao}{Perdig\~{a}o}

\acrodef{CNN}{convolutional neural network}
\acrodefplural{CNN}{convolutional neural networks}
\acrodef{DNN}{deep neural network}
\acrodefplural{DNN}{deep neural networks}
\acrodef{BVLOS}{beyond visual line of sight}
\acrodef{sUAV}{small uncrewed aerial vehicle}
\acrodefplural{sUAV}{small uncrewed aerial vehicles}
\newacroindefinite{sUAV}{an}{a}
\acrodef{UAV}{uncrewed aerial vehicle}
\acrodefplural{UAV}{uncrewed aerial vehicles}
\acrodef{CFD}{computational fluid dynamics}
\acrodef{RANS}{Reynolds-averaged Navier--Stokes}
\acrodef{LES}{large eddy simulation}
\newacroindefinite{LES}{an}{a}
\acrodef{NWP}{numerical weather prediction}
\newacroindefinite{NWP}{an}{a}
\acrodef{RMSE}{root mean squared error}
\newacroindefinite{RMS}{an}{a}
\acrodef{MSE}{mean squared error}
\newacroindefinite{MSE}{an}{a}
\acrodef{DEM}{digital elevation map}
\acrodef{TKE}{turbulence kinetic energy}
\acrodef{AoA}{angle of attack}
\newacroindefinite{AoA}{an}{an}
\acrodef{AoS}{angle of sideslip}
\newacroindefinite{AoS}{an}{an}
\acrodef{VAE}{variational auto-encoder}
\acrodef{GPR}{Gaussian process regression}
\acrodef{FPR}{flight path reconstruction}
\acrodef{IMU}{inertial measurement unit}
\newacroindefinite{IMU}{an}{an}
\acrodef{AMSL}{above mean sea level}
\acrodef{GNSS}{global navigation satellite system}
\acrodef{EZG}{EasyGlider}
\acrodef{WFLO}{wind farm layout optimization}
\acrodef{PDE}{partial differential equation}

\title{\LARGE \bf
WindSeer: Real-time volumetric wind prediction over complex terrain aboard a small UAV
}

\author{Florian Achermann$^{1}$, Thomas Stastny$^{1, 2}$, Bogdan Danciu$^{1}$, Andrey Kolobov$^{3}$, Jen Jen Chung$^{1, 4}$ \\ Roland Siegwart$^{1}$, Nicholas Lawrance$^{1, 5}$
\thanks{$^{1}$ Autonomous Systems Lab, ETH Z\"urich, Z\"urich 8092, Switzerland {\tt \footnotesize \{acfloria, danciub, rsiegwart\}@ethz.ch}}%
\thanks{$^2$ Auterion, Edenstrasse 20, Z\"urich 8045, Switzerland, { \tt\footnotesize thomas.stastny@auterion.com}}%
\thanks{$^3$ Microsoft Research, One Microsoft Way, Redmond 98052, USA, { \tt\footnotesize akolobov@microsoft.com}}%
\thanks{$^4$ School of Electrical Engineering and Computer Science, The University of Queensland, Staff House Road, QLD 4072, Australia, { \tt\footnotesize jenjen.chung@uq.edu.au}}%
\thanks{$^5$ Robotic Perception and Autonomy, CSIRO Data61, QLD 4069, Australia, { \tt\footnotesize nicholas.lawrance@csiro.au}}%
}

\begin{document}

\maketitle
\thispagestyle{empty}
\pagestyle{empty}

\begin{abstract}
Real-time high-resolution wind predictions are beneficial for various applications including safe manned and unmanned aviation. Current weather models require too much compute and lack the necessary predictive capabilities as they are valid only at the scale of multiple kilometers and hours  -- much lower spatial and temporal resolutions than these applications require. Our work, for the first time, demonstrates the ability to predict low-altitude wind in real-time on limited-compute devices, from only sparse measurement data. We train a neural network, \wcnn, using only \emph{synthetic} data from computational fluid dynamics simulations and show that it can successfully predict \emph{real} wind fields over terrain with known topography from just a few noisy and spatially clustered wind measurements. \wcnn\ can generate accurate predictions at different resolutions and domain sizes on previously unseen topography without retraining. We demonstrate that the model successfully predicts historical wind data collected by weather stations and wind measured onboard drones.
\end{abstract}

\begin{keywords}
    Wind Field Prediction, Machine Learning, Computational Fluid Dynamics, Uncrewed Aerial Vehicles, Sparse Data, Multi Resolution Prediction, Convolutional Neural Network, Low Altitude Wind
\end{keywords}

\section{Introduction}\label{introduction}
Accurate modelling of the wind is crucial for applications such as \ac{WFLO} or safe manned and unmanned aviation. The energy generated by wind turbines is proportional to the cubic power of the wind speed, thus micrositing turbines relies on accurate flow models~\cite{mattuella2016micrositing}. Adverse wind poses a challenge for manned aviation close to the ground at airports with challenging surrounding terrain, such as Madeira International Airport~\cite{belo2020madeira}. Finally, winds in mountainous regions can easily exceed \qty{10}{\meter\per\second}~\cite{ch_windatlas}, a speed comparable to the normal cruise speed of \acp{sUAV}~\cite{Oettershagen2015designatlantiksolar}, resulting in poor tracking of the planned flight path~\cite{Stastny2019guidance}. If \iac{sUAV} knew the wind in advance, path planning could avoid areas of unfavorable winds and high turbulence~\cite{chakrabarty2013windplanning}.

Chaotic fluid-dynamic effects due to local steep terrain result in large spatial variations of wind around complex terrain~\cite{buizza2002chaos}. Grid-based models thus require sufficiently high spatial resolution. \Ac{NWP} can accurately model relatively large-scale wind patterns with resolutions on the order of kilometers \cite{voudouri2018cosmo}. \Acf{CFD} simulations generate high-resolution wind flows around terrain at smaller scales, on the order of meters or less, but require knowing well-defined boundary conditions reflecting the overall weather situation~\cite{berg2011bolund, bechmann2011bolund}. Both simulation-based methods numerically solve the underlying system of \acp{PDE}, thus are computationally expensive (compute times in the order of hours) and do not provide real-time wind field estimates. Onsite wind measurements either with a Doppler Lidar~\cite{Vasiljevic2016lidar} or measurement masts~\cite{berg2011bolund, bechmann2011bolund, taylor1983askervein, taylor1987askervein, fernando2019perdigao} provide real-time wind information but with limited resolution and high setup cost.

AI-based methods have been used to accelerate the computation of flow fields by assisting or replacing numerical \ac{PDE}-based solvers in different settings. Examples include modelling fluid flows for visual rendering~\cite{xie2018tempoGAN, byungsoo2019} and replacing \ac{CFD} simulations in aerodynamic shape optimization~\cite{ribeiro2020deepcfd, bhatnagar2019prediction, umetani2018, baque2018geodesic, le2022towards}. However, these models rely on privileged information, such as boundary conditions and consider much simpler geometries compared to the topography of complex terrain. Super-resolution flow analysis closely aligns with our approach, yet previous studies in this field assume complete, uniform coverage of measurements over the entire region~\cite{guemes2022super, fukami2021global, fukami2023super, yang2023super}. They either exclusively investigate two-dimensional flows~\cite{guemes2022super, fukami2021global, fukami2023super} or require dense low-resolution data~\cite{guemes2022super, fukami2023super, yang2023super}. Various \ac{DNN}-based approaches demonstrated weather prediction at a global scale, essentially replacing \ac{NWP} with much faster compute time in the order of seconds~\cite{wandel2020nnfluid, pathak2022fourcastnet, bi2022pangu, lam2022graphcast}. But the resolution of these models, on the order of kilometers, is too low to accurately model the wind around complex terrain.

\begin{figure}
    \centering
    \includegraphics[width=1.0\textwidth]{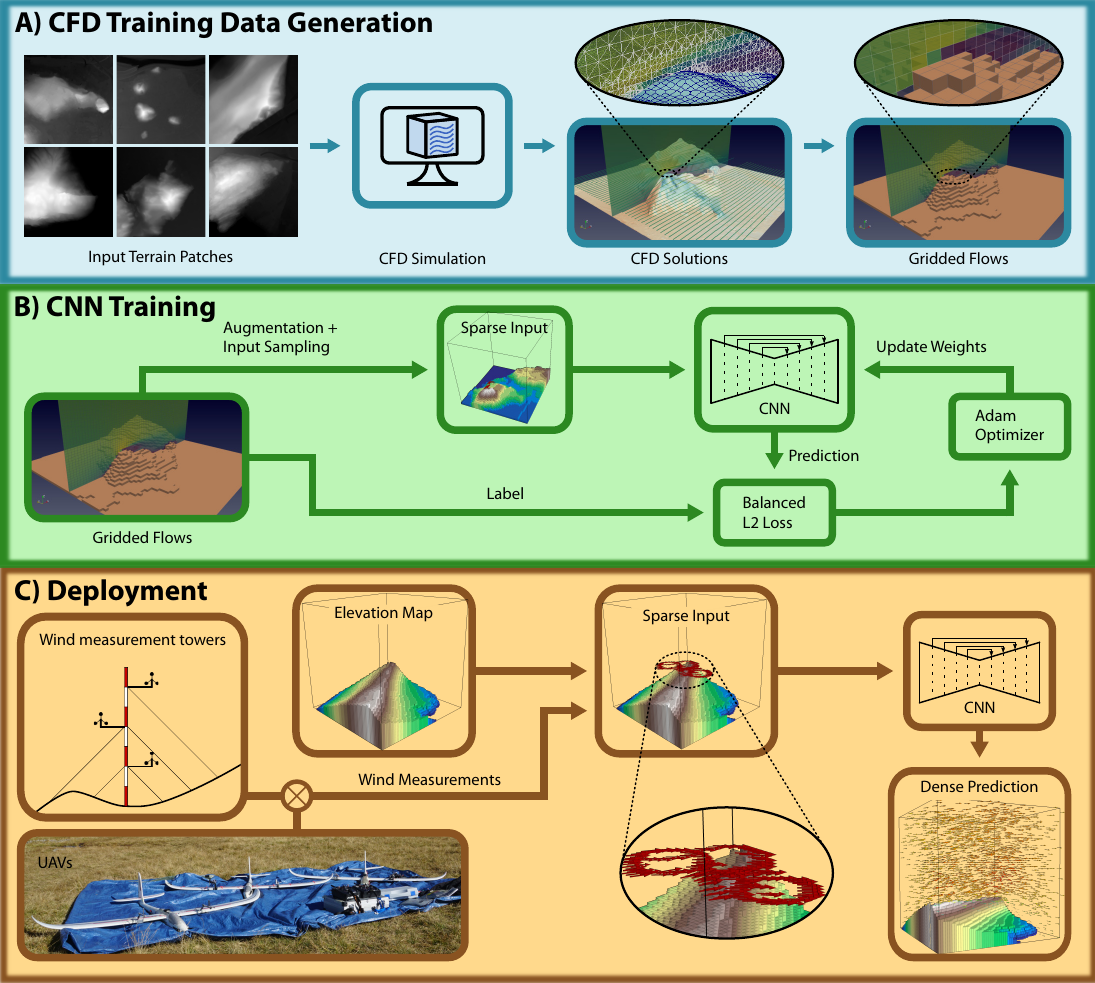}
    \caption[Pipeline Overview]{\textbf{Overview of the wind prediction pipeline}. (\textbf{A}) First we generate labelled flows utilizing a CFD simulation. (\textbf{B}) Then \wcnn\ is trained with measurements along randomly sampled piecewise linear trajectories to predict the dense flow. (\textbf{C}) During deployment the wind estimates from the UAV or wind measurement towers together with the known topography serve as the input to \wcnn.}
    \label{fig:overview}
\end{figure}

In this work, we present \wcnn, an approach for predicting the wind and turbulence in real-time at meter-scale based on the topography and sparse, noisy wind observations without needing bulky specialized equipment or assuming access to privileged information. \wcnn's ability to predict \emph{real} wind stems from its encoder-decoder \acf{CNN} architecture trained \emph{offline} using \emph{synthetic} flow data generated by computationally expensive steady-state \ac{RANS} \ac{CFD} simulations. The \ac{CFD} simulations are run offline over real terrain patches that are available from web services~\cite{usgs2021usmap, swisstopo2021swissmap}. The novelty of our method is in training \wcnn\ to produce CFD-like predictions from only sparse, in-situ observations -- \emph{without requiring privileged information} such as global boundary conditions. Access to these boundary conditions would be equivalent to measuring the wind along the full boundary of the prediction region, which is simply not available in real-time, nor at the required meter-scale. An overview of our wind prediction pipeline is presented in Fig.~\ref{fig:overview}. We evaluated \wcnn\ in a series of experiments, whose results show \wcnn's ability to make accurate dense wind and turbulence predictions based on local noisy wind measurements, across a wide range of spatial resolutions without retraining the model.

\clearpage

\section{Results}\label{results}
We evaluated \wcnn\ in a sequence of increasingly challenging experiments to demonstrate its real-time wind prediction capabilities:
\begin{enumerate}
    \item We demonstrated on held-out \ac{CFD}-simulated flows that \wcnn{} is expressive enough to represent the complex flow patterns around real terrain. We analysed several network training approaches on this dataset as it provides dense labels with a wide array of topographies.
    \item We demonstrated the ability of \wcnn{} to predict real wind data using measurements gathered from masts as part of large-scale measurement campaigns over different terrains across Europe~\cite{berg2011bolund, bechmann2011bolund, taylor1983askervein, taylor1987askervein, fernando2019perdigao}, thereby validating both the complete pipeline and our approach of using \ac{CFD} as a teacher model. These datasets offer good spatial coverage with measurement from different flow regions.
    \item On data from several multi-\ac{sUAV} flights over mountainous terrain we illustrate \wcnn's ability to predict the wind when using noisy onboard wind measurements as input. These input measurements were subject to high noise due to the uncertainty of the estimated \ac{sUAV}'s pose and errors in the airflow sensing, all of which complicated the prediction problem faced by \wcnn.
    \item Finally, we showed \wcnn's real-time prediction capability on flight-grade hardware.
\end{enumerate}

\subsection{\wcnn\ model}
\wcnn\ is a \acf{CNN} with four-channel input. The input is composed of a binary mask indicating cells containing input measurements, the terrain model stored as a distance field, and the sparse horizontal wind speed measurements (two channels). \emph{Note that vertical wind speed measurements are not an input to the model}, since weather stations typically measure only the horizontal wind and the vertical wind is not observable with a standard fixed-wing \ac{sUAV} sensor set. In Appendix~\ref{sec:app_ablation_study} we show empirically that adding vertical wind as an input, if it were available, has a limited impact on prediction quality. The percentage of observed cells in the input data varies across experiments ranging from \qty{3.5e-6}{\percent} to \qty{0.19}{\percent}, thus \wcnn\ always operates on highly sparse observations.

The four-channel output of \wcnn\ has the same spatial dimension and resolution as the input. The first three channels contain the three-dimensional wind prediction ($W_x, W_y, W_z$) and the fourth channel contains the \ac{TKE} prediction --- a metric for the strength of the turbulent velocity fluctuations in the wind field that is proportional to the sum of the variances in each dimension. Thus, the prediction contains properties ($W_z$, \ac{TKE}) that are not available as input measurements.

\subsection{Experiment group 1: Predicting \ac{CFD}-simulated flows}
We evaluated \wcnn\ on held-out \ac{CFD}-simulated flows. The dense label data allowed for evaluation over the full domain, thus evaluating the influence of different measurement locations as well as qualitatively characterizing the prediction quality.

\paragraph{Experiment setup}
We used \ac{CFD}-simulated flows over previously unobserved terrains and sampled the input measurements and noise from the same distributions as observed during training.

\begin{figure} [h!]
    \centering
    \includegraphics[width=1.0\textwidth]{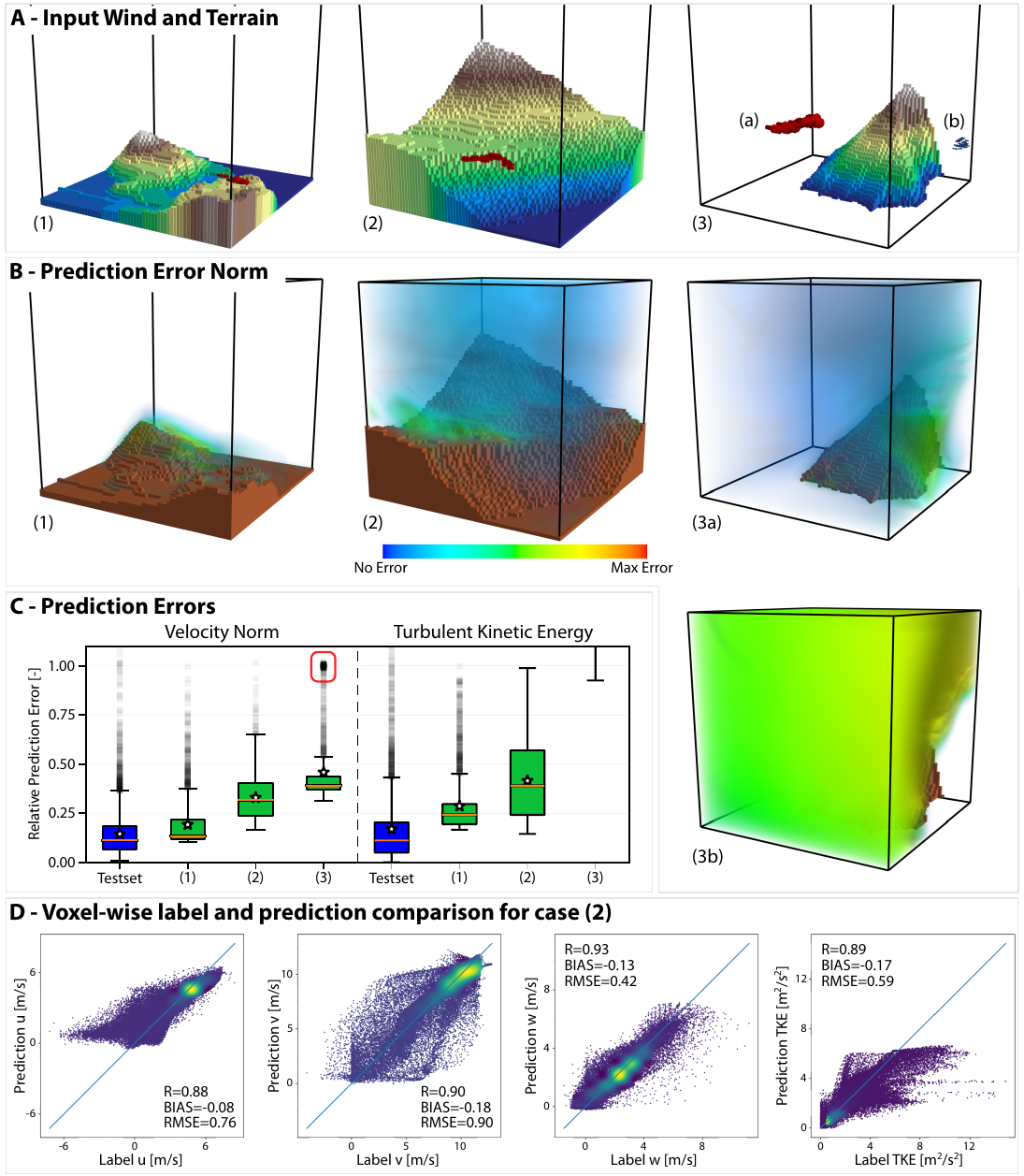}
    \caption[CFD Results]{\textbf{CFD experiment}. \textbf{A}) Terrain and input wind measurements (red arrows) with their respective prediction error (\textbf{B}). High prediction errors can be observed close to the ground or on the lee side of the terrain. \textbf{C}) Wind and turbulence prediction performance on the CFD dataset over the full test set (blue) and with 2000 random trajectories for the three different terrains shown in A. While most of the terrains result in uni-modal error distributions (1,2), more complex ones can have a second mode for samples from a complex flow region, indicated by the red box in (3). \textbf{D}) Density scatter plots comparing the label and the predictions for each predicted property using the terrain and input pair presented in A) (2).}
    \label{fig:cfd_results}
\end{figure}

\paragraph{Prediction performance}
Three terrain and input pairs together with the prediction error cloud are shown in Fig.~\ref{fig:cfd_results} A) and B). While the highest prediction errors occurred either close to the ground or on the lee side of the terrain, this trend is mitigated by the fact that, due to practical considerations such as payload configuration and safety, the operating altitude for \acp{sUAV} is typically over \qty{50}{m} above ground level~\cite{Oettershagen2017atlantiksolar, Ellis2021survey} and the wind turbine hub heights are typically higher than \qty{80}{\meter} above ground~\cite{lantz2019windtowerheight}. The distribution of the average normalized prediction errors over all non-terrain cells over the full test set is displayed in Fig.~\ref{fig:cfd_results} C) in blue. When scoring the network output only above an altitude of \qty{46}{m} the median relative velocity norm error reduces from \qty{14.5}{\percent} to \qty{11.5}{\percent} and the median relative \ac{TKE} error from \qty{11.2}{\percent} to \qty{8.3}{\percent}. The high correlation values in the voxel-wise comparison of the prediction to the \ac{CFD} labels shown in Fig.~\ref{fig:cfd_results} D) show that \wcnn{} can qualitatively capture the different flow regimes. The bias (average error) is close to zero for all channels and the \ac{RMSE} is also low compared to the overall magnitude. Together, this underlines \wcnn's prediction quality. The scatter density plots for the terrains presented in Fig.~\ref{fig:cfd_results} A) (1) and (3) are available in Extended Data Figure~\ref{fig:app_scatter_plots}.

\paragraph{Error analysis}
We evaluated three individual terrains in more detail (Fig.~\ref{fig:cfd_results} A)) to assess the sensitivity of the prediction quality to the sampled input data locations. For each terrain we randomly sampled 2000 trajectories and evaluated the prediction error (Fig.~\ref{fig:cfd_results} C) green). No noise or bias was added to the input data in this experiment to focus solely on the impact of the trajectory location. Whenever the input data was sampled in regions where the model could not predict the prevailing flow well, e.g. the lee side of the hill (Fig.~\ref{fig:cfd_results} A) (3b)), \wcnn{} performed poorly. If such a region is large enough, multi-modal error distributions can be observed as seen in Fig.~\ref{fig:cfd_results} C) (3) where prediction failed for input wind samples on the right (lee) side of the hill.

The \ac{CFD}-simulated flows are computed on a finer grid close to terrain to account for the high spatial variation of the flow in these regions. Interpolating the flow to the coarser fixed-size grid of \wcnn{}, as visible in Fig.~\ref{fig:overview} A), leads to a loss of information and flow artifacts close to the ground. Such confounding factors make learning these low-altitude flows especially challenging, offering an explanation for the performance difference between predicting the low-altitude and high-altitude winds. However, the experiments with the real wind data show that the wind close to the ground can be accurately modelled if the grid resolution is increased.\\

\subsection{Experiment group 2: Evaluation on wind measurement campaign datasets}
\label{sec:experiments_measurement_campaigns}
We evaluated \wcnn\ on real wind data available from three published measurement campaigns. The long measurement periods allow filtering out short-term effects such as wind gusts and reduces the overall measurement noise, enabling an evaluation of \wcnn\ with clean input wind measurements over larger-scale domains.

\begin{figure}
    \centering
    \includegraphics[width=0.9\textwidth]{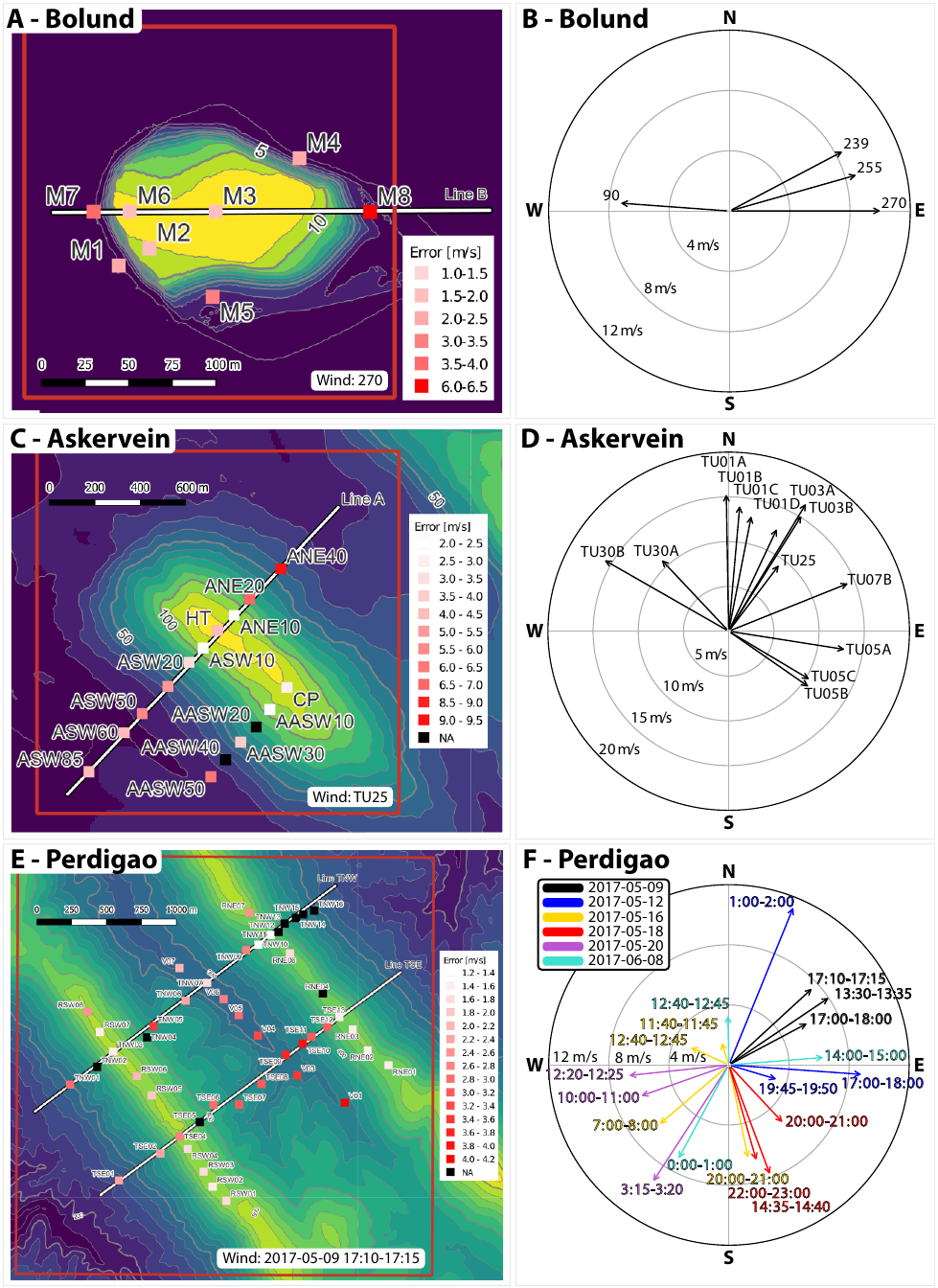}
    \caption[Intro Measurement Campaigns]{\textbf{Measurement campaigns experiment}. The mast locations and elevation maps for the Bolund (\textbf{A}), Askervein (\textbf{C}), and \Perdigao{} (\textbf{E}) campaigns. The tower positions are colored by the average prediction error when using that specific mast as the input to predict the wind. In the Askervein and \Perdigao{} case some masts did not provide a valid measurement for that experiment. (\textbf{B}, \textbf{D}, \textbf{F}) show the wind directions for the different experiments for each terrain.}
    \label{fig:meas_campaign_cases}
\end{figure}

\paragraph{Experiment setup}
The in-situ wind and \ac{TKE} measurements were collected via wind velocity sensor suites (sonic or cup anemometers) mounted on masts providing data from \qty{2}{m} to \qty{100}{m} altitude above ground level~\cite{fernando2019perdigao}. For each terrain, varying wind flow directions and magnitudes are available from different measurement periods (Fig.~\ref{fig:meas_campaign_cases}). The terrain in these campaigns varies in complexity and size -- from the \qty{11}{m} high Bolund hill (Fig.~\ref{fig:meas_campaign_cases} A))~\cite{berg2011bolund, bechmann2011bolund} to the gently-sloped \qty{116}{m} high Askervein hill (Fig.~\ref{fig:meas_campaign_cases} C))~\cite{taylor1983askervein, taylor1987askervein}. Both Bolund and Askervein have limited vegetation while the \Perdigao{} region in Portugal represents the most complex test case with two lightly-forested parallel ridges roughly \qty{300}{m} high (Fig.~\ref{fig:meas_campaign_cases} E)).

The varying geometric extents of the sites together with the low altitude wind measurements required a larger grid size ($384\times384\times192$ instead of $64^3$ cells) paired with higher resolution to obtain meaningful predictions. Accordingly, the grid resolutions were increased $2\times$, $4\times$, and $30\times$ for the \Perdigao{}, Askervein, and Bolund terrains, respectively. These changes in the prediction grid were enabled by the fully convolutional architecture of \wcnn\ and the distance field representation of the terrain that indirectly provides the cell size and resulted in around $100\times$ sparser input data compared to the training density.

We compared \wcnn\ against an averaging baseline (AVG) that assumes the wind and \ac{TKE} are constant and predicts the average of all measurements over the full domain. This is a widely accepted assumption for \ac{sUAV} flights~\cite{Stastny2019guidance, jayaweera2022wind, coombes2019windcoverage}, moreover, the state of the art in planning large-scale missions relies on \ac{NWP} forecasts that remain constant at the spatial resolution of the mission~\cite{Oettershagen2019metpass}. Each method predicted the wind based on the measurements from a single mast, yielding an ensemble of predictions for each wind case, while measurements from the remaining masts were used to validate the predictions.

\paragraph{Prediction performance}
For each wind case we averaged the metrics over the ensemble of predictions and report the absolute errors of the wind magnitude, vertical wind, and \ac{TKE} in Tab.~\ref{tab:error_real_wind}. In most cases \wcnn\ outperformed the averaging baseline (AVG). In the other cases, the wind direction usually aligned with the ridge/terrain, causing only small variations across the measurements of the different masts. The average prediction errors of \wcnn\ are \qty{17}{\percent}, \qty{43}{\percent}, and \qty{39}{\percent} lower than the baseline for the velocity magnitude, vertical wind and \ac{TKE} respectively.

To assess whether \wcnn\ can predict flow trends well, we also report the correlation between the wind predictions and measurements in Tab.~\ref{tab:error_real_wind}. The correlation for the AVG baseline is undefined, due to the constant, location-invariant wind prediction. Averaged over all cases, \wcnn\ yielded strong positive correlations for all metrics suggesting that it was able to predict the observed trends well, such as up- and downdrafts, high wind and turbulence, thus providing a valuable contribution to planning safer and more efficient \ac{sUAV} trajectories or \ac{WFLO}.

\begin{figure} [h!]
    \centering
    \includegraphics[width=\textwidth]{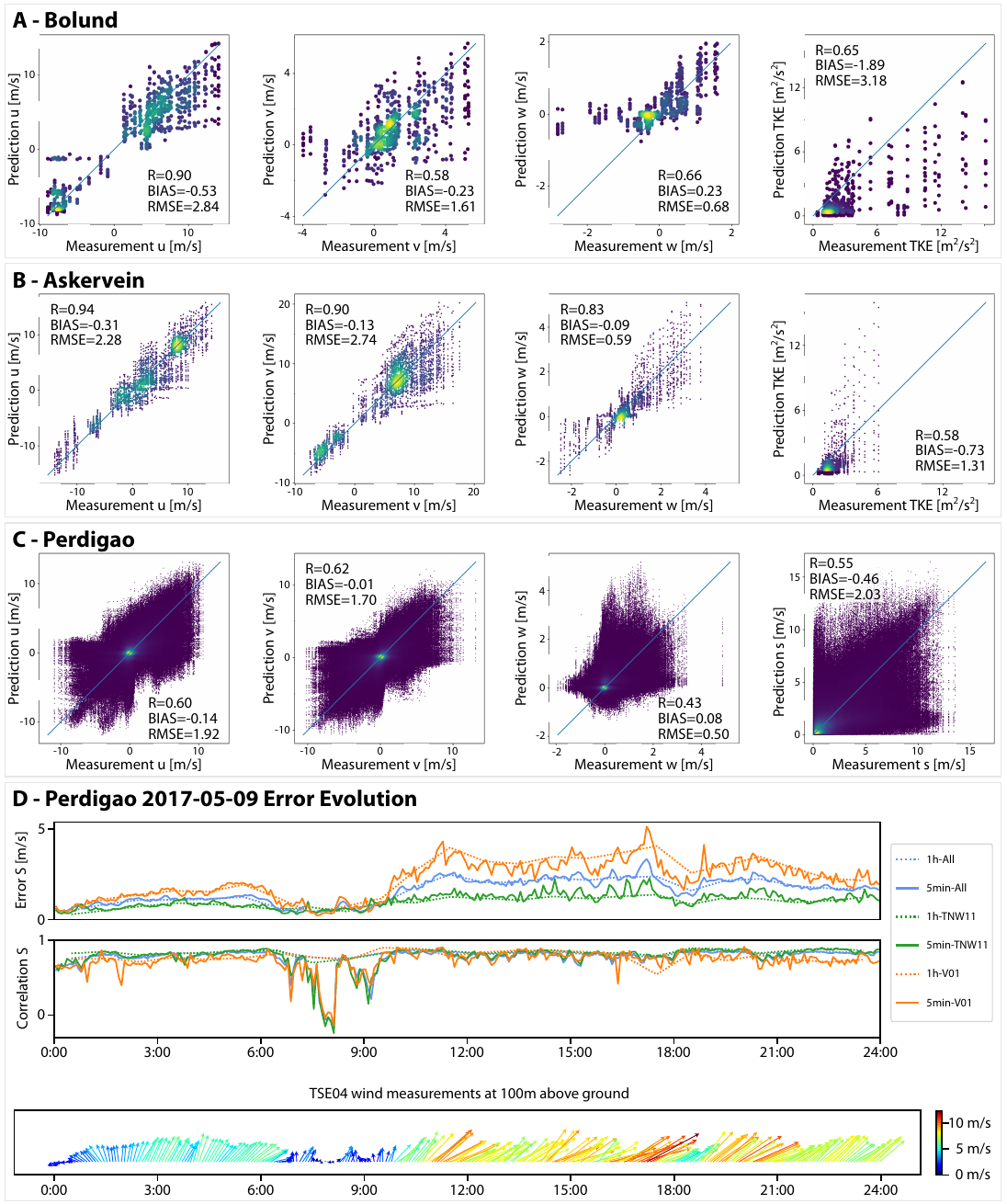}
    \caption[Scatter Plots Measurement Campaigns]{\textbf{Measurement campaign results}. Measured wind compared to the predictions aggregated over all predictions for the Bolund (\textbf{A}, 32 predictions [4 experiments, 8 masts]), Askervein (\textbf{B}, 182 predictions [13 experiments, 14 masts]), and \Perdigao{} (\textbf{C}, 9120 predictions [240 experiments, 38 masts]) campaigns. In \textbf{D} the evolution of the prediction error and correlation of the wind norm S for \wcnn{} (WS) using the 5 minute and 1 hour averaged data together with the measurements from the TSE04 tower as a reference are shown. We show the results aggregated over all the 38 predictions using the different masts as input and the scores using only the TNW11 and V01 tower data for the prediction.}
    \label{fig:meas_campaign_scatter}
\end{figure}

In Fig.~\ref{fig:meas_campaign_scatter} A-C we compare the measurements to the \wcnn{} predictions. Analogous to the previous experiment we use the wind measurements from one mast as input resulting in 32 predictions for the Bolund [8 masts and 4 wind cases], 182 predictions for the Askervein [14 masts and 13 wind cases], and 9120 predictions for the \Perdigao{} campaigns [38 masts and 240 wind cases (one hour averaged data for ten different days\footnote{2017-05-09,  2017-05-11, 2017-05-12, 2017-05-16, 2017-05-18, 2017-05-20, 2017-05-26, 2017-06-02, 2017-06-03, 2017-06-08})]. Overall, the low bias and \ac{RMSE} together with the high correlation demonstrate that \wcnn{} handles different wind conditions well. The best predictions are obtained for the simpler Askervein terrain and the errors grow with increasing terrain complexity. In general the model under-predicts the downdrafts in the wind fields for the Bolund and Perdigao cases (Fig.~\ref{fig:meas_campaign_scatter} A and C) but the Askervein experiment (Fig.~\ref{fig:meas_campaign_scatter} B) shows that \wcnn{} is capable of representing strong downdrafts.

For the Bolund hill a study of different \ac{CFD} simulation was conducted and the speed up errors for one wind direction (\qty{239}{\degree}) are reported~\cite{bechmann2011bolund}. The average absolute speedup prediction error for \wcnn\ is \qty{20.3}{\percent} while the best performing \ac{RANS}-\ac{CFD} models achieve an error of \qty{15}{\percent} but with runtimes in the order of hours. The models with a runtime of less than \qty{15}{\min} have errors of \qtyrange{26.5}{32.4}{\percent}, comparable to our averaging baseline with error \qty{33.5}{\percent}.

\begin{table}
    \centering
    {\tabulinesep=0.4mm
    \begin{tabu}{l l | c c c | c c c | c c c}
    \hline
     &  & \multicolumn{3}{c|}{S $[m/s]$} & \multicolumn{3}{c|}{W $[m/s]$} & \multicolumn{3}{c}{TKE $[m^2/s^2]$} \\
     &  & AVG & \multicolumn{2}{c|}{WindSeer} & AVG & \multicolumn{2}{c|}{WindSeer} & AVG & \multicolumn{2}{c}{WindSeer} \\
    Terrain & Case & $\epsilon$ & $\epsilon$ & $\rho$ & $\epsilon$ & $\epsilon$ & $\rho$ & $\epsilon$ & $\epsilon$ & $\rho$ \\
    \hline
    \multirow{4}{*}{Bolund} & 90 & 1.80 & \textbf{1.58} & 0.72 & 0.85 & \textbf{0.58} & 0.50 & 1.91 & \textbf{1.33} & 0.58\\
    & 239 & 2.80 & \textbf{2.50} & 0.68 & 0.66 & \textbf{0.34} & 0.76 & 2.67 & \textbf{1.68} & 0.86\\ 
    & 255 & 3.24 & \textbf{2.47} & 0.82 & 0.85 & \textbf{0.44} & 0.72 & 3.43 & \textbf{2.12} & 0.82\\ 
    & 270 & 3.77 & \textbf{2.79} & 0.85 & 0.95 & \textbf{0.51} & 0.78 & 5.14 & \textbf{3.43} & 0.73\\
    \hline
    \multirow{13}{*}{Askervein} & TU25 & 2.58 & \textbf{2.39} & 0.65 & 1.10 & \textbf{0.37} & 0.90 & 0.61 & \textbf{0.41} & 0.89\\ 
    & TU30A & 1.14 & \textbf{0.98} & 0.62 & 0.41 & \textbf{0.26} & 0.58 & 1.38 & \textbf{0.72} & 0.42\\
    & TU30B & 1.80 & \textbf{1.46} & 0.73 & 0.51 & \textbf{0.41} & 0.64 & 2.82 & \textbf{1.40} & 0.23\\ 
    & TU01A & 3.26 & \textbf{2.83} & 0.77 & 1.41 & \textbf{0.52} & 0.91 & 1.89 & \textbf{1.06} & 0.85\\ 
    & TU01B & 3.24 & \textbf{2.74} & 0.79 & 1.37 & \textbf{0.48} & 0.92 & 1.64 & \textbf{0.98} & 0.87\\ 
    & TU01C & 3.55 & \textbf{3.08} & 0.78 & 1.23 & \textbf{0.45} & 0.92 & 1.17 & \textbf{0.72} & 0.90\\
    & TU01D & 4.21 & \textbf{3.71} & 0.79 & 1.26 & \textbf{0.47} & 0.93 & 1.62 & \textbf{1.17} & 0.93\\
    & TU03A & 5.29 & \textbf{4.70} & 0.78 & 1.74 & \textbf{0.64} & 0.93 & 2.04 & \textbf{1.31} & 0.98\\ 
    & TU03B & 4.90 & \textbf{4.41} & 0.77 & 1.54 & \textbf{0.54} & 0.92 & 1.82 & \textbf{1.21} & 0.90\\ 
    & TU05A & 1.91 & \textbf{1.89} & 0.62 & 0.76 & \textbf{0.31} & 0.89 & 1.73 & \textbf{0.99} & 0.40\\ 
    & TU05B & 1.18 & \textbf{1.00} & 0.79 & 0.31 & \textbf{0.26} & 0.48 & 1.40 & \textbf{0.63} & 0.04\\
    & TU05C & 0.93 & \textbf{0.93} & 0.66 & 0.34 & \textbf{0.25} & 0.58 & 1.09 & \textbf{0.43} & 0.14\\
    & TU07B & 3.42 & \textbf{3.27} & 0.70 & 1.59 & \textbf{0.49} & 0.90 & 2.44 & \textbf{1.85} & 0.40\\
    \hline
    \multirow{3}{*}{\shortstack[l]{\Perdigao{}\\ 2017-05-09}} & 13:30-13:35 & 2.91 & \textbf{2.27} & 0.82 & 0.85 & \textbf{0.57} & 0.53 & - & - & - \\
    & 17:10:17:15 & 4.41 & \textbf{3.33} & 0.48 & 1.22 & \textbf{0.88} & 0.35 & - & - & - \\
    & 17:00-18:00 & 3.06 & \textbf{2.37} & 0.77 & 0.80 & \textbf{0.56} & 0.50 & - & - & - \\
    \hline
    \multirow{3}{*}{\shortstack[l]{\Perdigao{}\\ 2017-05-12}} & 01:00-02:00 & 2.82 & \textbf{2.31} & 0.76 & 0.52 & \textbf{0.42}  & 0.45 & - & - & - \\ 
    & 17:00-18:00 & 2.76 & \textbf{2.23} & 0.81 & 0.70 & \textbf{0.57} & 0.57 & - & - & - \\
    & 19:45-19:50 & 1.15 & \textbf{0.90} & 0.71 & 0.21 & \textbf{0.16} & 0.57 & - & - & - \\
    \hline
    \multirow{4}{*}{\shortstack[l]{\Perdigao{}\\ 2017-05-16}} & 07:00-08:00 & 1.58 & \textbf{1.16} & 0.65 & 0.27 & \textbf{0.27} & 0.67 & - & - & - \\
    & 11:40-11:45 & 0.85 & \textbf{0.77} & 0.51 & 0.33 & \textbf{0.27} & 0.22 & - & - & - \\
    & 12:40-12:45 & 0.86 & \textbf{0.75} & 0.48 & 0.29 & \textbf{0.22} & 0.30 & - & - & - \\
    & 20:00-21:00 & 1.35 & \textbf{0.97} & 0.68 & \textbf{0.22} & 0.31 & 0.21 & - & - & - \\
    \hline
    \multirow{3}{*}{\shortstack[l]{\Perdigao{}\\ 2017-05-18}} & 14:35-14:40 & 2.14 & \textbf{1.82} & 0.70 & 0.41 & \textbf{0.36} & 0.33 & - & - & - \\
    & 20:00-21:00 & 1.54 & \textbf{1.08} & 0.84 & \textbf{0.17} & 0.19 & 0.12 & - & - & - \\
    & 22:00-23:00 & 1.52 & \textbf{1.13} & 0.74 & 0.17 & \textbf{0.17} & 0.42 & - & - & - \\
    \hline
    \multirow{3}{*}{\shortstack[l]{\Perdigao{}\\ 2017-05-20}} & 03:15-03:20 & 3.59 & \textbf{2.63} & 0.61 & \textbf{0.41} & 0.55 & 0.30 & - & - & - \\
    & 10:00-11:00 & 2.20 & \textbf{1.84} & 0.71 & 0.51 & \textbf{0.42} & 0.46 & - & - & - \\
    & 12:20-12:25 & 1.93 & \textbf{1.71} & 0.66 & 0.58 & \textbf{0.43} & 0.45 & - & - & - \\
    \hline
    \multirow{3}{*}{\shortstack[l]{\Perdigao{}\\ 2017-06-08}} & 00:00-01:00 & 2.68 & \textbf{1.87} & 0.69 & \textbf{0.35} & 0.38 & 0.44 & - & - & - \\
    & 12:40-12:45 & 1.20 & \textbf{0.90} & 0.77 & 0.31 & \textbf{0.22} & 0.46 & - & - & - \\
    & 14:00-15:00 & 2.49 & \textbf{1.77} & 0.82 & 0.74 & \textbf{0.42} & 0.58 & - & - & - \\
    \hline
    \multicolumn{2}{l|}{All campaigns} & 2.50 & \textbf{2.07} & 0.71 & 0.72 & \textbf{0.41} & 0.59 & 2.05 & \textbf{1.26} & 0.64 \\
    \hline
    \hline
    \multirow{3}{*}{\shortstack[l]{Chasseral}} & Flight 1 & \textbf{0.61} & 0.85 & 0.33 & 0.54 & \textbf{0.38} & 0.95 & - & - & - \\
    & Flight 2 & \textbf{0.57} & 0.77 & 0.48 & 0.50 & \textbf{0.32} & 0.80 & - & - & - \\
    & Flight 3 & \textbf{0.53} & 0.71 & 0.37 & 0.49 & \textbf{0.30} & 0.84 & - & - & - \\
    \hline
    Oberalp & Flight 1 & \textbf{0.56} & 0.82 & -0.46 & 0.54 & \textbf{0.33} & 0.78 & - & - & - \\
    \hline
    Gotthard & Flight 1 & \textbf{0.99} & 1.13 & 0.33 & 0.21 & \textbf{0.18} & 0.88 & - & - & - \\
    \hline
    \multicolumn{2}{l|}{All flights} & \textbf{0.65} & 0.86 & 0.21 & 0.46 & \textbf{0.30} & 0.85 & - & - & - \\
    \hline
    \end{tabu}}
    \caption[Real World Results]{\textbf{Real world wind results}. Absolute prediction errors ($\epsilon$) and correlation between the measurements and predictions ($\rho$) for the velocity magnitude (S), vertical wind component (W), and turbulence kinetic energy (TKE) on the measurement campaign datasets and flight experiments} of \wcnn{} compared to the averaging baseline (AVG). Note that the AVG baseline does not offer correlations as it produces a constant signal.
    \label{tab:error_real_wind}
    \vspace{-2mm}
\end{table}

\paragraph{Error Analysis}
The \ac{RANS} \ac{CFD} simulations compute the time-averaged solution of the Navier-Stokes equation, thus \wcnn{} is trained to operate on these static flows. However, in the real world, wind changes constantly. The \Perdigao{} campaign provides measurements as 5-minute averages allowing us to compare the performance of \wcnn{} operating with high- and low-frequency data as shown in Fig.~\ref{fig:meas_campaign_scatter} D. Averaged over a day, the performance is consistent across the two time windows with slightly better results using the hourly averages. One exception is the large difference in the correlation scores observed between 06:00-10:00, a time period characterized by low wind magnitudes and rapidly changing wind direction as evident from the TSE04 tower measurements. These dynamic conditions do not match the \ac{RANS} \ac{CFD} simulation offering an explanation for the poorer prediction performance on the 5-minute averaged data compared to the hourly averages that filter out these unsteady flow features.

In Fig.~\ref{fig:meas_campaign_cases} the masts are colored by the averaged wind magnitude error of the \wcnn\ prediction when using the measurements from that respective mast as inputs. Lighter colors indicate input measurement mast locations that yielded more accurate predictions. In Fig.~\ref{fig:meas_campaign_scatter} D we show the prediction errors and correlation using the TNW11 and V01 tower data in addition to the averaged errors over all tower predictions. Consistent with the findings from our previous experiments, using measurements from the hill top or the upwind side generally resulted in lower-error predictions than measurements from the lee side, nevertheless, the correlation is consistently high.

\begin{figure} [h!]
    \centering
    \includegraphics[width=1.0\textwidth]{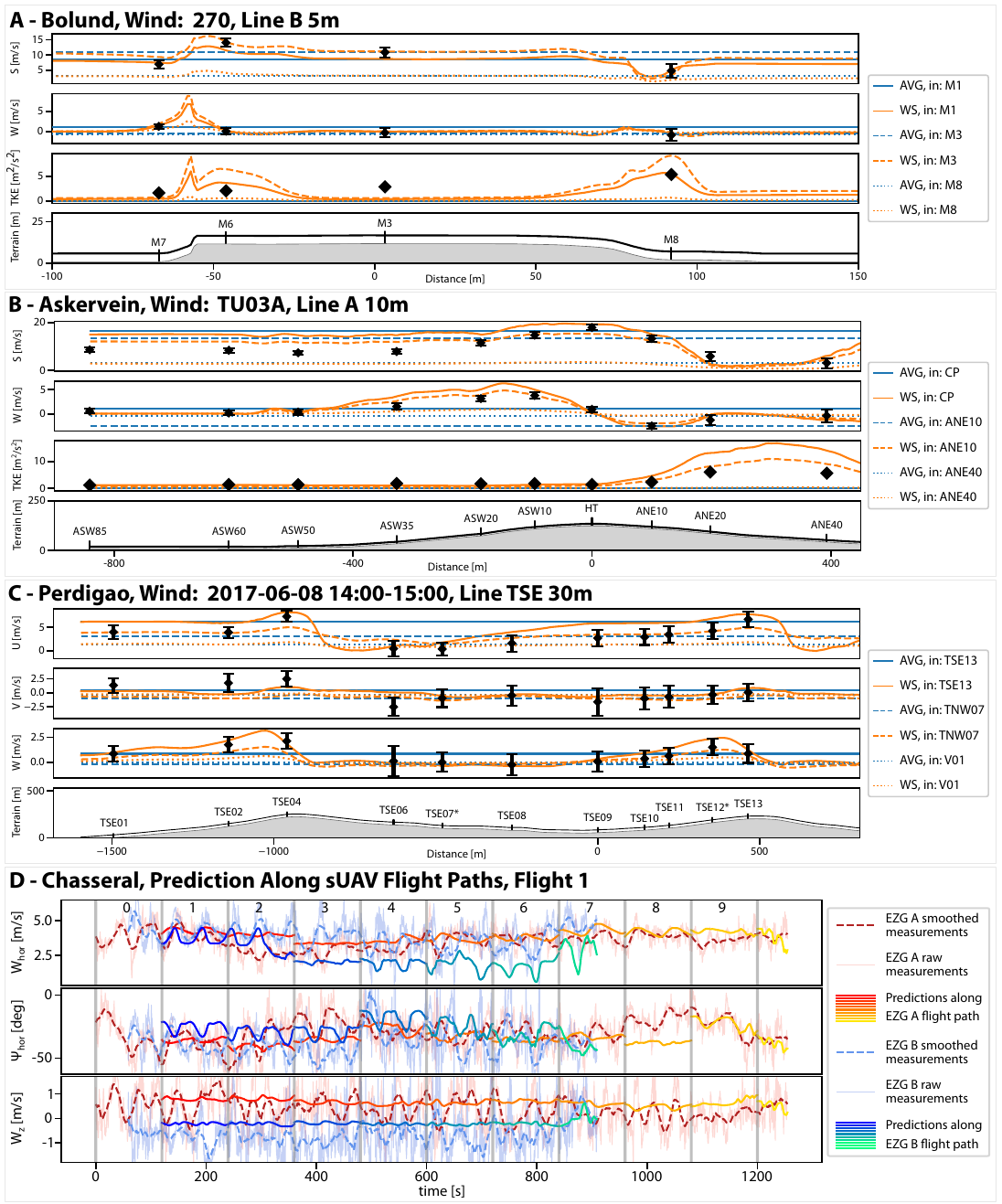}
    \caption[Results Measurement Campaigns]{\textbf{Measurement campaigns prediction line and sUAV predictions along the path}. \textbf{(A-C)} Predictions and measurements along characteristic lines with a constant height for each experiment with the baseline averaging method (AVG) and \wcnn{} (WS). Three predictions using different input masts are shown for each model and experiment. The asterisk * indicates that no measurement was available for that respective mast at the queried height and the closest one was picked. The uncertainty of the measurements is displayed by the standard deviation of the raw high-rate data. \textbf{(D)} The predictions from EZG A along the flight paths from EZG A and B for the first Chasseral flight.}
    \label{fig:meas_campaign_result}
\end{figure}

\paragraph{Wind field trends along straight lines}
In each campaign the masts were arranged along multiple straight lines enabling us to qualitatively assess whether the models could capture expected flow trends along these lines. We selected cases where the wind and the line direction are parallel, as the measurements show higher variation in these scenarios, and present the \wcnn\ and baseline predictions in Fig.~\ref{fig:meas_campaign_result}. Refer to Fig.~\ref{fig:meas_campaign_cases} for the wind direction and the mast locations. For each method and case we show three predictions using the measurements from different masts as the input. The error bars for wind speed measurements report the 1$\sigma$ uncertainty.

\wcnn\ successfully predicted the speed changes and up-/downdrafts unless the measurement tower was located on the lee side of a hill, e.g.  Bolund: M8, Askervein: ANE40, \Perdigao{}: V01/TNW07. Wherever \ac{TKE} measurements were available, \wcnn{} predicted \ac{TKE} trends well. \wcnn\ struggles to predict flow patterns not observed during training, such as a lee side rotor which occurs when flow detaches on the downwind side and causes a recirculating pattern (Appendix~\ref{sec:app_wind_char}). Such a case is shown in Extended Data Figure \ref{fig:meas_campaign_result_extended_data}.

\subsection{Experiment group 3: Predicting the wind along \ac{sUAV} trajectories}
We evaluated \wcnn\ on noisy, local wind measurements collected by multiple fixed-wing \acp{sUAV}. This mirrors the targeted use case of \wcnn\ and challenges it with high input noise levels due to real sensor noise and short term wind effects such as gusts or turbulence.

\paragraph{Experiment setup}
We flew multiple fixed-wing \acp{sUAV} simultaneously in the Swiss Jura (Chasseral) and the Swiss Alps (Oberalppass and Gotthardpass). The flight plans for each \ac{sUAV} consisted of multiple circular loiter patterns. We generated the sparse input from the point-cloud of measurements to \wcnn\ by binning the observations along the flight path from one \ac{sUAV} into the discretized prediction grid and averaging all measurements falling in a single cell. We used the native training grid size and resolution with the grid center at the first observed \ac{sUAV} position estimate.

\paragraph{Prediction performance}
We first evaluated \wcnn\ on time-averaged data, similar to the previous experiment group with the static masts, to reduce the noise and sensitivity to sensor calibration on the input measurements. We generated this data by averaging the measurements over one loiter pattern to a single wind measurement. The observations from the different \acp{sUAV} and multiple loiters enabled us to compute the average prediction error and the correlation for each flight experiment. We present the error metrics for all flights in Tab.~\ref{tab:error_real_wind}. For the wind magnitude, the baseline outperforms \wcnn\, which also only yields a slightly positive correlation averaged over all flights. Nevertheless, the high correlations and significantly lower errors for the vertical wind compared to the baseline indicate that \wcnn\ can better predict the locations of dangerous downdrafts, as well as favourable updraft regions, based solely on wind measurements taken from one \ac{sUAV}.

We further evaluated \wcnn\ in a sequential time-windowed manner using the wind data from a \qty{120}{\second} window to predict the wind along the flight trajectory for the next window. The corresponding predictions for the first Chasseral flight are displayed in Fig.~\ref{fig:meas_campaign_result}. Note that the prediction between successive windows can be discontinuous due to the different measurements used by \wcnn. In Extended Data Figures \ref{fig:flight_tests} and \ref{fig:flight_tests_appendix} we provide the predictions for the other flights as well as flight paths and example predictions. The model was able to accurately predict the magnitude difference in the vertical wind between the two \acp{sUAV} for all Chasseral flights. It slightly under-predicted the downwind on the lee side for the validation \ac{sUAV}, which can be explained by the generally worse performance of the models on the lee-side wind predictions, as shown in the previous experiments.

\paragraph{Error Analysis}
Although the measurements were averaged over time the noise due to wind gusts and measurement errors was comparable to the variation of the measured wind magnitude as outlined in Appendix~\ref{sec:app_airflow_sensing}. Thus, the averaging already offered a good baseline prediction of the wind magnitude. As the wind on the vertical axis varied much more between the different \acp{sUAV} and throughout a flight, \wcnn\ could predict these variations and outperform the baseline.

The Oberalppass and Gotthardpass are especially challenging prediction terrains, as they exhibit large altitude changes (\qty{1500}{\meter}) due to valleys and peaks within \qty{4}{\kilo\meter} of our flight locations. High surrounding peaks can cause high gust levels, explaining the high variation in the measured wind. Furthermore, terrain outside the prediction area can significantly influence the wind features observed in the valley. In contrast, our \ac{CFD} simulation setup for generating the \wcnn\ training data was limited to well-defined inflow conditions and a domain size of \qtyproduct{1.5 x 1.5}{\kilo\metre}, which did not allow simulating the flow over multiple large scale mountains or ridges. Thus, during training \wcnn\ did not observe such complex wind flows, explaining the performance difference between the Chasseral and Oberalppass/Gotthardpass flights.

\subsection{\wcnn\ inference time}
We evaluated prediction times of \wcnn\ on an NVIDIA Jetson Orin AGX, a light-weight and low-power single-board computer, to show real-time performance on \ac{sUAV} flight-grade hardware. The average inference times over 100 runs on a $64^3$ and $384\times384\times192$ prediction domain were \qty[separate-uncertainty-units = single, separate-uncertainty=true]{0.021(2)}{\second} and \qty[separate-uncertainty-units = single, separate-uncertainty=true]{3.577(15)}{\second} respectively. Mixed-precision inference reduced the inference times to \qty[separate-uncertainty-units = single, separate-uncertainty=true]{0.021(5)}{\second} and \qty[separate-uncertainty-units = single, separate-uncertainty=true]{1.700(5)}{\second}. These inference times show that \wcnn\ is capable of low-latency wind predictions over large domains with limited compute to quickly recalculate predictions in response to new measurements.

\clearpage

\section{Discussion}\label{discussion}
In this work, we have proposed an approach to train a \ac{CNN}, \wcnn, for predicting low-altitude wind and \ac{TKE} around complex terrain in real-time based on sparse and noisy wind measurements and known topography. We trained \wcnn\ solely on simulated \ac{RANS} \ac{CFD} flows over terrain patches from Switzerland and evaluated it on held-back \ac{CFD} data and real wind measurements. In the first experiment on previously unobserved \ac{CFD} solutions we demonstrated that \wcnn\ is capable of replicating the dense flows based on sparse and noisy observations with high accuracy (median relative error below \qty{10}{\percent}).

In the next experiments we demonstrated zero-shot sim-to-real transfer by evaluating \wcnn\ on real wind measurements without retraining. On the historic measurement campaign datasets we showed that \wcnn\ was able to reconstruct real wind flows of different scales, at up to 30 times higher resolution than the training data. This corresponds to larger prediction domains containing up to 108 times more cells than those used for network training and to much sparser input data compared to training. The distance field representation enabled this multi-resolution property of \wcnn\ as it provided a sense of scale to the model.

Finally, the performance of \wcnn{} on the flight data is comparable to the average baseline assumption, with less accurate predictions of the horizontal wind but superior performance when predicting the vertical wind, which is the key factor when assessing the safety and efficiency of flight plans. The flight data exhibits much higher measurement noise due to the noise of the low-cost sensors used to estimate the \ac{sUAV} pose and wind. Therefore, until better wind sensing is available on \acp{sUAV}, we envision the onboard deployment of \wcnn\ by using the weather data from nearby weather stations.

Previous work has shown the ability of \acp{DNN} to predict fluid flows for well-defined geometries paired with well-known inflow conditions~\cite{ribeiro2020deepcfd, bhatnagar2019prediction, umetani2018, Kashefi2021fluid, Zhang2021flowaroundstructure}. We demonstrated the capability of \acp{DNN} to work with sparse and noisy input data on realistic complex terrain. This enables real-time wind prediction using data that is feasible to obtain aboard \iac{sUAV}. The sparsity of the input data (\qty{0.19}{\percent} down to \qty{3.5e-6}{\percent}) exceeds previous research of sparse-to-dense \acp{DNN} that usually assume denser data around \qty{0.75}{\percent}~\cite{Fangchang2018sparse2dense,Jaritz2018sparse}, \qty{0.2}{\percent}~\cite{Kaiyue2020depthcompletion}, or \qty{6.5e-3}{\percent}~\cite{Huang2018rmsnet}. In these previous examples the sparse input data was distributed over the whole prediction domain while in our case we showed \wcnn\ still performs well even if the samples are located within a spatially constrained sub-region.\\

\subsection{Limitations and Future Work}
\paragraph{Training domain}
In our data generation pipeline we restricted the \ac{CFD} simulation domain to \qtyproduct{1.5 x 1.5}{\kilo\metre} based on the initial assumption that the large scale \ac{NWP} would be used in the network input. This domain size restricted the terrain to mostly contain one single major geographical feature such as a mountain or a ridge. Therefore the current \ac{CFD} training data does not contain samples that include wind phenomena such as lee-side rotors, which arise in the presence of multiple mountains/ridges. A larger simulation domain in the order of \qtyproduct{10 x 10}{\kilo\metre} could allow a better representation of such complex flows and possibly increase the wind prediction performance for complex terrains inside mountain ranges. A biased sampling strategy when composing the input could also help to solve the sample imbalance problem by exposing the network to more examples from the lee-side flow regime during training.

\paragraph{Temporal wind variation}
Changing the \ac{CFD} simulation from a time-averaged \ac{RANS} solution to a time-varying model such as \ac{LES} or a mesoscale weather prediction, such as the WRF model~\cite{skamarock2019wrf}, could have multiple advantages. First, \iac{DNN} could be trained using the time-varying wind data to construct the input but still predict the time-averaged solution. This could result in the model learning wind gust characteristics and thus increase robustness to noisy wind estimates from the \ac{sUAV}. Second, a model could be trained to predict the time-varying flow representing wind gusts and short term weather evolution in the predicted wind. However, whether the information from the noisy measurements are sufficient to uniquely determine the flow state still needs to be carefully analyzed. Depending on the sparsity of the data there are likely to be multiple possible flow solutions matching the observations. Further, time varying methods like \ac{LES} require significantly more computational resources than \ac{RANS} solvers, further increasing the cost of generating training data~\cite{blocken2014}.

\paragraph{Fluid flow assumptions}
Currently the \ac{CFD} simulations used to train \wcnn{} model the air as an incompressible fluid with uniform temperature. By including temperature differences in the compressible fluid and terrain the \ac{CFD} simulation could model complex flow phenomena such as thermals~\cite{akos-bb10}, updrafts caused by temperature variations on the ground, or mountain waves~\cite{Durran1986mountainwave, Chang2018simulatingmountawaves}, which are large scale oscillations of the wind direction and magnitude behind large ridges. However, simulating these phenomena would require far more input data and ultimately we would still need to verify whether these simulations provide realistic flows that reflect the true airflow characteristics.

\paragraph{Wind Estimation}
Our current wind sensing setup onboard the \ac{sUAV} is prone to calibration errors and noise resulting in relatively large wind estimation errors. Alternative sensors, such as a five hole probe~\cite{Sankaralingam2021fhp} paired with an improved calibration procedure or better sensor placement could improve the wind estimates.


\section{Methods}\label{methods}
\subsection{Overview}
We developed a pipeline (Fig.~\ref{fig:overview}) to train and deploy \iac{CNN}, \wcnn, that predicts the dense wind and turbulence around complex terrain. The network training consists of two steps: First we generated a dataset of dense flows over terrain patches from Switzerland using \iac{RANS} \ac{CFD} solver (Fig.~\ref{fig:overview} A)). We then trained \wcnn\ using the label flows to simulate local wind measurements along randomly generated piecewise-linear trajectories, robustifying the predictions by adding noise to the measurements along the trajectories (Fig.~\ref{fig:overview} B)). The trained \wcnn\ was evaluated on (i) held back \ac{CFD}-simulated flows on previously unobserved terrains, (ii) real wind data gathered in measurement campaigns~\cite{berg2011bolund, bechmann2011bolund, taylor1983askervein, taylor1987askervein, fernando2019perdigao}, and (iii) real wind data measured by multiple \acp{sUAV} around mountainous terrain.\\

\subsection{\ac{CFD} wind data}
We generated flow data over real terrain patches with a pipeline based on the open source solver OpenFOAM~\cite{weller1998foam, jasak2007} with the steady-state \ac{RANS} model and the popular $k-\epsilon$ two-equation turbulence closure~\cite{launder1974}. The automated pipeline ingests terrain patches and outputs the time-averaged flow solutions for multiple wind speeds~\cite{achermann2019windprediction}. We extracted 563 terrain patches each with an extent of \qtyproduct{1.5 x 1.5}{\kilo\metre} from the GeoVite service, which provides access to the swissALTI3D \ac{DEM} for Swiss researchers, with a lateral resolution of \qty{0.5}{m}\footnote{\url{https://geovite.ethz.ch/}, recently also available on ArcGIS: \newline \url{https://elevation.arcgis.com/arcgis/rest/services/WorldElevation/Terrain/ImageServer}}. The terrain patches exhibit at least one side with near-constant elevation allowing us to simulate a formed boundary layer flow (logarithmic profile) entering into the domain from that face. Some terrains allowed for multiple flow directions leading to 866 terrain/flow direction pairs. The vertical extent of the simulation domain was three times the height difference of the terrain with a lower bound of \qty{1100}{\metre} minimizing the boundary effects on the flow. Each case was simulated with up to 15 different wind speeds if the automatic meshing succeeded, resulting in 7361 executed \ac{CFD} runs. We initialized the subsequent simulation for the higher wind speed cases with the previous solution to speed up computation. Only solutions that met a required optimization tolerance were accepted as fully converged solutions, which was the case in \qty{92.9}{\percent} of the runs. We enhanced our dataset with one zero-velocity flow for each terrain that had at least one converged \ac{CFD} simulation, resulting in a total of 7285 flows.

The \ac{CFD} solutions are computed on an automatically generated irregular mesh with OpenFOAM's SnappyHexMesh utility. We resampled each case up to a height of \qty{1100}{\metre} to a regular \numproduct{91 x 91 x 96} grid resulting in a resolution of \qty{16.5}{\metre} horizontally and \qty{11.5}{\metre} vertically (Fig.~\ref{fig:overview} A)).\\

\subsection{Data augmentation}
Generating \ac{CFD} flows is a computationally and labor-intensive task. For reference, our 7361 \ac{CFD} runs required \qty{9168}{\hour} CPU compute time (\qty{782}{\hour} creating the meshes and \qty{8386}{\hour} solving the flow, average compute time: \qty{1.25}{\hour}). Unfortunately, deep networks are notoriously data-hungry and, for a complex modeling problem such as wind prediction, would typically require orders of magnitude more training data to achieve good performance. In computer vision, image augmentation methods are widely used when training deep \acp{CNN}~\cite{Shorten2019augmentation}. These methods aim to improve the quality and size of the datasets when only limited data is available to prevent the networks from over-fitting. In this work, we showed, for the first time, that geometric transformations can be applied to \ac{CFD} flows to augment the \wcnn\ training data.

We randomize the locations of terrain features and flow directions by generating $64^3$ subdomains sampled from each full \numproduct{91 x 91 x 96} grid. The subdomains are constructed by sampling from a range of rotations and origin translation offsets inside the full domain. In a first step the horizontal shift and a rotation around the $z$-axis are sampled from bounded uniform distributions to ensure that the shifted and rotated $64^2$ subdomain is fully contained within the full $91^2$ domain. Then in a second step the vertical shift is sampled from a triangle distribution with lower limit and mode of 0 and an upper limit of 32. Smaller vertical offsets are favoured to focus on the complex flow regions closer to the terrain. The flow data is linearly interpolated to the coordinates of the subdomain grid, which is the same spatial resolution as the full grid. \\

\subsection{Input and label composition}
The input to \wcnn\ consists of four volumetric channels, one of which corresponds to the terrain encoding $T$ created as a Euclidean distance transform with zeros inside the terrain. That representation propagates the terrain information over the full domain and even allows us to include terrain features outside of the domain if they are accounted for in the distance field calculation. The remaining three channels include the sparse and noisy horizontal wind measurements ($U_{x,in},U_{y,in}$) and a binary mask $B$ indicating cells containing measurements. Previous work has shown the value of providing binary input masks to \acp{CNN} handling sparse input data~\cite{Kohler2014mask, Uhrig2017sparsemask}.

Measurements from weather stations or realistic flight scenarios only cover a small percentage of the prediction volume along a connected path, e.g. a \qty{30}{\second} flight segment with our \ac{sUAV} covers approximately 20 cells at the default grid resolution. Consequently, for a practical onboard wind prediction scenario, we expect the available input wind data to be very sparse and thus construct our network input to reflect this sparsity. We create the input based on the augmented dense flow by creating a mask and then selecting the measurements based on the mask. We emulate the characteristics of \iac{sUAV} flight path by filling the mask along sequential randomly-selected piecewise linear segments with a length of 3 to 500 cells.

Noise is added to the sampled wind data in order to account for fluctuations in the wind and sensor errors that are not captured by the \ac{RANS} \ac{CFD} simulations. Two types of disturbance are added, white Gaussian noise (sampled i.i.d. at each measurement from $\mathcal{N}\left(0, \sigma_g^2\right)$) and measurement bias (sampled from $\mathcal{U}\left(-0.1, 0.1\right)$ and applied to all measurements). The first has the purpose of simulating noise due to sensor measurements~\cite{Nirmal2016noise}, while the latter simulates the effects of sensor miscalibration. The standard deviation for the Gaussian noise $\sigma_g$ itself is drawn from a uniform distribution: $\sigma_g \sim \mathcal{U}\left(0, 0.1\right)$, simulating different noise levels. All the noise values are scaled with the mean wind velocity for each sample in the training set to have coherent noise levels from low to high velocity samples. Note that noise is only added to the training inputs and not to the CFD ground truth labels used to compute the network training losses.

The sparse input implies that for most cells in the input wind velocity channels ($U_{x,in},U_{y,in}$) the values are undefined since they do not contain a measurement. We test and evaluate two approaches to filling the missing information. The first na\"{i}ve approach simply places zeros in all voxels without a measurement. This results in large gradients of the input for high magnitude wind. The second approach uses the per-channel-average of all measurements as the fill value resulting in a smoother input and propagating the information over the whole domain.

The labels are constructed by stacking the four volumetric channels corresponding to the three-dimensional predicted velocity ($U_{x,out}, U_{y,out}, U_{z,out}$) as well as the \ac{TKE} at each cell from the CFD ground truth flows.\\

\subsection{Model training}
The wind prediction model is an encoder-decoder \ac{CNN} with skip connections based on the U-Net architecture~\cite{ronneberger2015u}. The \wcnn\ encoder is composed of single 3D convolutions with kernel size 3 and reflection padding to preserve the size. Using skip connections, the information at each depth is relayed to the decoder before utilizing a max-pooling layer with kernel size of 2 to down-sample the feature map. The original domain size is restored by pairing the information from the skip connection with a nearest-neighbor up-sampled feature map followed by two 3D convolutions with kernel size 4. This decoder structure removes checkerboard artifacts sometimes experienced when using a decoder-encoder \ac{CNN} \cite{odena2016deconvolution}. Each convolution, except the final one, is followed by the nonlinear ReLu layer with negative slope 0.1 \cite{maas2013relu}.

A majority of the cells in the wind speed input channels contain no measurements. Our first approach was to set the values of all these cells to zero. However, this resulted in a network overfitting to the number of observed cells and did not generalize to larger domains and different resolutions as demonstrated in Appendix~\ref{sec:app_ablation_study}. In the end, we ended up filling the unobserved cells with the average of all measurements per channel which results in a smoother input and helps to propagate the information over the full domain.

A scaled version of the \ac{MSE} loss is applied to train the model balancing the loss $L(\cdot)$ between the samples and channels:
\begin{equation}
    L\left(X,Y,N\right) = \frac{1}{N}\sum_{c}
    \left(\frac{X_{c} - Y_{c}}{\hat{Y}_{c}}\right)^2,
\end{equation}
where $X$ is the network prediction, $Y$ the label flow, $\hat{Y}_{c}$ the label average per channel of the non-terrain cells, and $N$ the number of non-terrain cells. Normalizing the error by the average label value balances the loss for flows of different magnitudes. Without accounting for the number of terrain cells in the loss, a sample with a high ratio of terrain cells would not contribute much to the overall loss. Thus, scaling according to $N$ prevents these cases from being underrepresented in the training.

The model is trained using the Adam optimizer \cite{kingma2015adam} for 3000 epochs. The initial learning rate of $1.0 \times 10^{-5}$ is quartered every 700 epochs.\\

\subsection{Measurement campaign datasets}
\label{sec:method_measurement_campaigns}
Each of the three measurement campaign datasets that we used for evaluation are publicly available but require some preprocessing to enable direct comparison with our wind prediction outputs. We convert the data from the different file formats for each measurement campaign to the same gridded format that we use to store the \ac{CFD} solutions. Each experiment provides terrain data as well as wind measurements collected using static masts equipped with airflow sensors at various heights. The terrain is discretized by querying the raw data using bilinear interpolation in the center of the respective cell. The location of each measurement is converted into the cell coordinates.

\wcnn\ predicts the wind using the measurements from one mast. The measurements are filled into the corresponding cell and averaged in case of multiple measurements in one cell. The predictions, which are obtained with trilinear interpolation at the sensing locations, are then compared to the measured wind. \acp{CNN} allow for variable input sizes, a trait we exploit to predict at a higher spatial resolution for domains with smaller length scales (see Bolund Hill below) at an increased domain size of $384\times384\times192$ cells. Since the terrain is represented as a Euclidean distance field, this gives \wcnn\ a sense of the grid resolution and thus the scale of the flow, enabling us to predict the wind at different scales.

The error bars for Fig.~\ref{fig:meas_campaign_result} are calculated using error propagation from the standard deviations of each axis if not reported for the magnitude~\cite{Ku1966propagation}.

\paragraph{Bolund hill}
The data for the Bolund hill experiment containing time-averaged wind velocities and \ac{TKE} measurements is publicly available\footnote{\url{https://www.bolund.vindenergi.dtu.dk/blind\_comparison}}. As Bolund hill exhibits only a small elevation change of \qty{11}{\meter}, the default prediction resolution is not sufficient to account for its near-ground measurement locations. As mentioned above, we exploit the multi-scale property of \wcnn\ and increase the resolution of the prediction grid thirty-fold resulting in a domain $\qty{\sim 211}{\meter}\times\qty{\sim 211}{\meter}$ wide and \qty{\sim 73}{m} tall, giving a corresponding horizontal resolution of \qty{0.55}{\meter} and vertical resolution of \qty{0.38}{\meter}.

\paragraph{Askervein hill}
While a digitized version of the Askervein hill topography is available\footnote{\url{https://zenodo.org/record/4095052}} the wind and \ac{TKE} measurements had to be manually extracted from the field report \cite{taylor1983askervein}. We selected 13 runs measuring the turbulent wind, where the data from most towers is provided (in certain runs data is not reported for all towers). The measurements are averaged over one- to four-hour intervals with varying flow magnitudes and directions. The domain size of $\qty{1584}{\meter}\times\qty{1584}{\meter}$ wide and \qty{552}{m} tall results in a four-fold resolution increase.

\paragraph{\Perdigao{}}
The \Perdigao{} dataset consists of multiple measurement posts of different heights ranging from \qty{10}{m} up to \qty{100}{m} across the valley or along the ridges\footnote{\url{https://perdigao.fe.up.pt/}}. We used the five minute averages and tilt corrected measurements that were recorded throughout the measurement campaign and we consider data from six different days in our evaluation. The tower positions were not stored with sufficient precision in the dataset requiring us to manually correct the positions. We extracted the topography of the hills from the World Elevation Terrain layer provided by Esri using ArcGIS \footnote{\url{https://www.arcgis.com/home/item.html?id=58a541efc59545e6b7137f961d7de883}}. \Perdigao{} required the largest prediction domain size, $\qty{3168}{m}\times\qty{3168}{m}$ wide and \qty{1104}{m} tall, showcasing the wind prediction performance at double the original resolution.\\

\subsection{Inference time experiments setup}
We ran the inference time experiments on an Orin AGX\footnote{\url{https://www.nvidia.com/en-us/autonomous-machines/embedded-systems/jetson-orin/}}, a low power, light weight (\qty{623}{\gram} including the carrier board and heatsink) and small scale (\qtyproduct{105 x 105 x 60}{\milli\meter}) single-board computer that can be carried by a small scale \ac{sUAV}. We set up the Orin AGX with the Jetpack 5.1 software kit that includes CUDA 11.4 and cuDNN 8.6.0 and installed PyTorch 2.0. During the evaluation we ran the Orin in the maximum power mode (\qty{60}{\watt}) using all 12 CPU cores.

\subsection{\ac{sUAV} flight tests}
\label{method:flight_tests}
We used three Multiplex EasyGlider4 airframes equipped with the Pixhawk 4 autopilot~\cite{meier2011pixhawk} using the high quality ADIS16448 \ac{IMU} and the u-blox M9N GNSS module for autonomous navigation. We configured the main height source of our modified PX4 autopilot~\cite{meier2015px4} to the GPS height and use the barometric pressure as a fallback. An extension to the guidance law adjusting the airspeed ensured safety during strong wind conditions~\cite{Stastny2019guidance}. We used a custom designed pitot tube with the Sensirion SDP31 differential pressure sensor and Hall sensor airflow vanes to enable measuring the 3D wind vector. Refer to Section~\ref{sec:app_airflow_sensing} for more details about the airflow sensing setup and calibration procedure. We used a ground station computer with QGroundControl to control and navigate the \ac{sUAV}. While the default PX4 state estimator could be extended to estimate the 3D wind we opted for an offline \ac{FPR} pipeline using an iterated extended Kalman filter (see a similar problem definition in~\cite{multer1999fpr}). The offline \ac{FPR} pipeline allowed us to generate high quality estimates for validating our approach and to adjust the estimation pipeline post flight.

We gathered wind data from flights at three test sites in Switzerland. The first test site at Chasseral is one of the most topographically isolated mountains in Switzerland and is located in the Jura mountains (47\textdegree~07'~38''~N, 7\textdegree~02'~47''~E, \qty{1548}{\meter} \ac{AMSL}). The other test sites are located on the ridges of the Oberalppass (46\textdegree~39'~24''~N, 8\textdegree~40'~21''~E, \qty{2069}{\meter} \ac{AMSL}) and Gotthardpass (46\textdegree~34'~17''~N, 8\textdegree~33'~33''~E, \qty{1960}{\meter} \ac{AMSL}) in the Central Swiss Alps. These were chosen to evaluate the prediction performance for domains surrounded by complex terrain. The spatial constraints allowed for two \acp{sUAV} to simultaneously collect wind data at Oberalppass and Gotthardpass and three sUAVs at Chasseral. The \acp{sUAV} were flown simultaneously in circular loiter patterns with a radius of \qty{100}{m} leading to lateral separation between the planes of up to \qty{800}{m} and measurements in different flow regimes. We planned the flights based on \ac{NWP} forecasts ensuring good flight (no precipitation, fog or clouds) and stable wind conditions (wind magnitude below the cruise speed of \qty{10}{\meter\per\second}, direction and magnitude near-constant over multiple hours).

We use two modes to convert the raw wind estimates to the \wcnn\ input. In the first mode we generated the input by averaging the measurements over one loiter pattern to generate a single wind estimate. This time-averaged data, similar to the averaged data in the measurement campaigns with static masts, helped to reduce the noise and sensitivity to sensor calibration on the input measurements. The observations from the different \acp{sUAV} and multiple loiters enabled us to compute the average prediction error and the correlation for each flight experiment to see if the flow trends are well predicted. In the second mode the input is composed of the wind data from a \qty{120}{\second} window of one \ac{sUAV} to predict the wind along the flight path within the next \qty{120}{\second} window for both the input (itself) and the validation \acp{sUAV}. This sequential time-windowed setup allowed us to qualitatively evaluate the \wcnn\ performance along the flight paths. A high-resolution elevation map provided by SwissTopo~\cite{swisstopo2021swissmap} was used to construct the terrain for the WindSeer input.

\section*{Acknowledgments}
We would like to thank our drone pilots David Rohr, Jaeyoung Lim, Jonas Langenegger, Tizian Steiger, and Yves Allenspach for ensuring the safety of our drones during the experiments. The arduous task of setting up the drones and the respective sensing devices was supported by Himmet Kaplan, Jonas Langenegger, Michael Riner, Thomas Mantel, Tizian Steiger, and Yves Allenspach. Last we would like to thank Alexey Dosovitskiy, Debadeepta Dey, and René Ranftl for their valuable inputs throughout the project.

This project was funded, in part, by Microsoft Swiss Joint Research Center under Contract No. 2019-038 ``Project Altair: Infrared Vision and AI Decision-Making for Longer Drone Flights'', the Intel Network on Intelligent Systems, and by the ETH Research Grant AvalMapper ETH-10 20-1.

The model training and evaluations were performed on the ETH Zürich Euler computing cluster.

\section*{Declarations}

\subsection*{Competing interests}
The authors declare that they have no competing interests.

\subsection*{Availability of data and materials}
The synthetic \ac{CFD} training data, as well as the \ac{sUAV} flight measurements will be made available through our dataset server: \url{https://projects.asl.ethz.ch/datasets/doku.php?id=nmi_23_windseer}.

\subsection*{Code availability}
The code is available on github: \url{https://github.com/ethz-asl/WindSeer}

\bibliographystyle{IEEEtran}
\bibliography{sn-bibliography}

\clearpage

\begin{appendices}

\section{Wind characteristics terminology}
\label{sec:app_wind_char}
\begin{figure}
    \centering
    \includegraphics[width=0.85\textwidth]{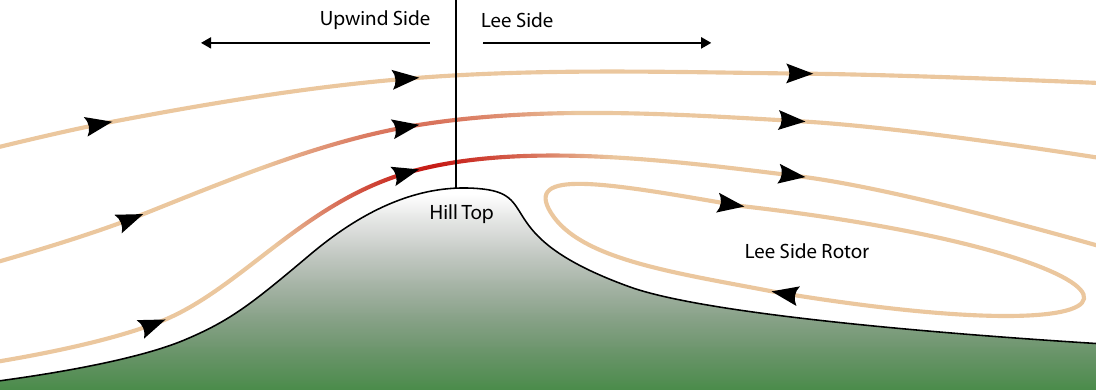}
    \caption[Flow over terrain]{Example of wind over a hill resulting in a lee side rotor.}
    \label{fig:flow_over_terrain}
\end{figure}

The complex wind around terrain has some typical flow regions and we want to establish common terms for some of these regions as shown in Fig.~\ref{fig:flow_over_terrain}. The side of the terrain where the wind direction points toward the hill is called the \textit{upwind} side since it typically exhibits mostly rising winds. At the highest point of the terrain (\textit{hill top}) the wind speeds up and higher wind magnitudes can be measured. The \textit{lee side} of the terrain/hill describes the region where the wind direction typically points away from the hill top. This region is highly turbulent and can form multiple modes with completely different characteristics. Under certain conditions the flow can follow the terrain downhill resulting in prevailing downwinds. In other conditions a \textit{lee side rotor} can form, where the wind follows a circular motion close to the terrain behind the hill. At a larger scale of multiple kilometers and under very specific conditions, mountain waves with multiple rotors can form~\cite{Durran1986mountainwave}. 

\section{\ac{NWP} data as \wcnn\ input}
\label{sec:app_nwp_input}
Our initial hypothesis was to train \wcnn\ based on the known high-resolution terrain and predictions from large scale \ac{NWP}. The Swiss COSMO 1 model provides predictions with a horizontal resolution of \qty{1.1}{\kilo\meter}~\cite{voudouri2018cosmo}. The elevation data used in the \ac{NWP} models, such as GLOBE~\cite{hastings1999global}, is an aggregation of available high-resolution terrain sources (usually the mean or median of the high-resolution data within one cell). This smoothed topography representation neglects smaller scale terrain features and therefore only provides meaningful results at a scale of multiple cells/kilometers.

\begin{figure} [b]
    \centering
    \includegraphics[width=0.95\textwidth]{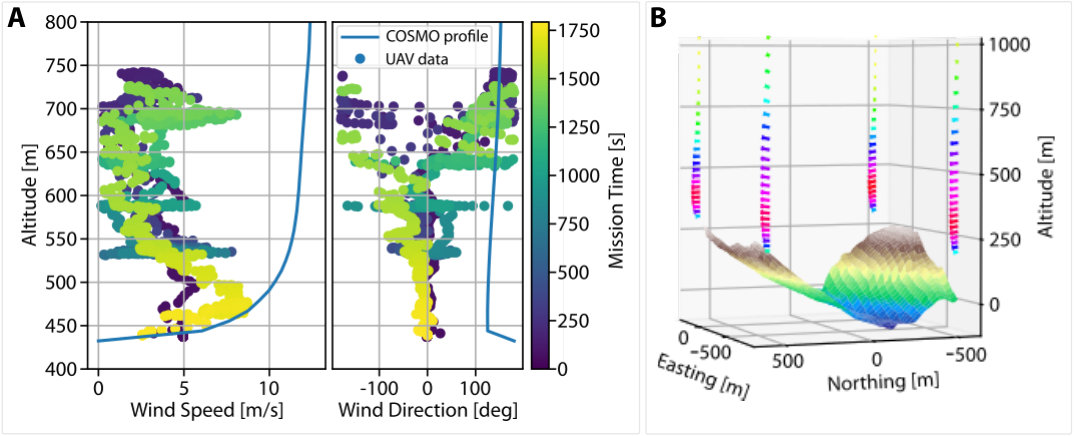}
    \caption[Flow over terrain]{(\textbf{A}) \ac{sUAV} wind measurements compared to the \ac{NWP} along different flight altitudes showing the mismatch in direction and wind speed. (\textbf{B}) The coarse terrain representation in the \ac{NWP} causes offsets in the prediction altitudes to the terrain.}
    \label{fig:nwp_comparison}
\end{figure}

We conducted a test flight to evaluate how well the \ac{NWP} of one cell matches the wind measured by \iac{sUAV}. We measured the wind at one grid point of the Swiss COSMO-1 model\footnote{\url{https://www.meteoschweiz.admin.ch/home/mess-und-prognosesysteme/warn-und-prognosesysteme/cosmo-prognosesysteme.html}} close to Flüelen (46\textdegree~53'~33''~N, 8\textdegree~36'~45''~E, \qty{436}{\meter} \ac{AMSL}). While Flüelen is located within the Swiss Alps, this particular test site is bordered on one side by a lake and surrounded by flat and smooth terrain within a \qty{1}{\kilo\meter} radius resulting in a good match between the \ac{NWP} terrain model and the high-resolution terrain. 

As visible in Fig.~\ref{fig:nwp_comparison}~A), the COSMO-1 \ac{NWP} poorly represents the \ac{sUAV} data for both the magnitude and wind direction. Obviously in different conditions the \ac{NWP} might fit the measurements better. However, this implies that, depending on the case, the \ac{NWP} may or may not be accurate. Thus, \wcnn\ needs another, more reliable source for its wind prediction data. In addition, the coarse representation of the terrain can result in large altitude offsets of the \ac{NWP} compared to the actual terrain in the presence of large elevation changes, see Fig.~\ref{fig:nwp_comparison}~B).

The \ac{NWP} data may provide supplemental information to \wcnn\ if used together with the sparse measurements. However, first the mapping between the \ac{NWP} data to the actual flow needs to be established. This would be a highly data-driven task, and if that connection is too noisy, \wcnn\ might learn to ignore the \ac{NWP} input data altogether.

\section{WindSeer ablation study}
\label{sec:app_ablation_study}
We evaluated the effect of varying certain hyperparameters in the training pipeline on the model performance on a test set of previously unobserved \ac{CFD} samples. The baseline model parameters are shown in Tab.~\ref{tab:baseline_hyperparameter}, note that these parameters are different from the finalized \wcnn\ version. We used the average error norm over all non-terrain cells averaged over all samples in the test set as our metric to compare the models.

\begin{table} [h]
    \centering
    \caption[Baseline hyperparameter set]{Baseline hyperparameter set used in the ablation study.}
    \begin{tabular}{l  l}
    \toprule
    \textbf{Hyperparameter} & \textbf{Value}\\
    \midrule
    learning rate       & $1.0 \times 10^{-5}$ \\
    learning rate decay & 0.25 every 700th epoch \\
    learning epochs & 1500 \\
    learning batch size & 35 \\
    max Gaussian noise std      & \qty{0}{\percent}\\
    max bias magnitude          & \qty{0}{\percent}\\
    trajectory min length       & 3 cells \\
    trajectory max length       & 50 cells \\
    model depth & 4 \\
    pooling method & strided convolution \\
    input no measurement value & mean \\
    input use $u_z$ & true \\
    \bottomrule
    \end{tabular}
    \label{tab:baseline_hyperparameter}
\end{table}

Models trained with different pooling methods (max-pooling (MP), average-pooling (AP), convolution with strides) perform comparably with a slight edge for the pooling methods over the convolution with strides (\qty{1.1}{\percent} error reduction). The model using only the horizontal wind measurement (NUZ) outperforms the baseline (BL) model, which uses the vertical measurements as well, by \qty{2.6}{\percent}. We also varied the input trajectory lengths up to a length of 500 cells (LT). Networks trained on longer trajectories perform \qty{13.6}{\percent} better even if they are evaluated exclusively on short trajectories with lengths of up to 50 cells.

\begin{figure} [h!]
    \centering
    \includegraphics[width=0.6\textwidth]{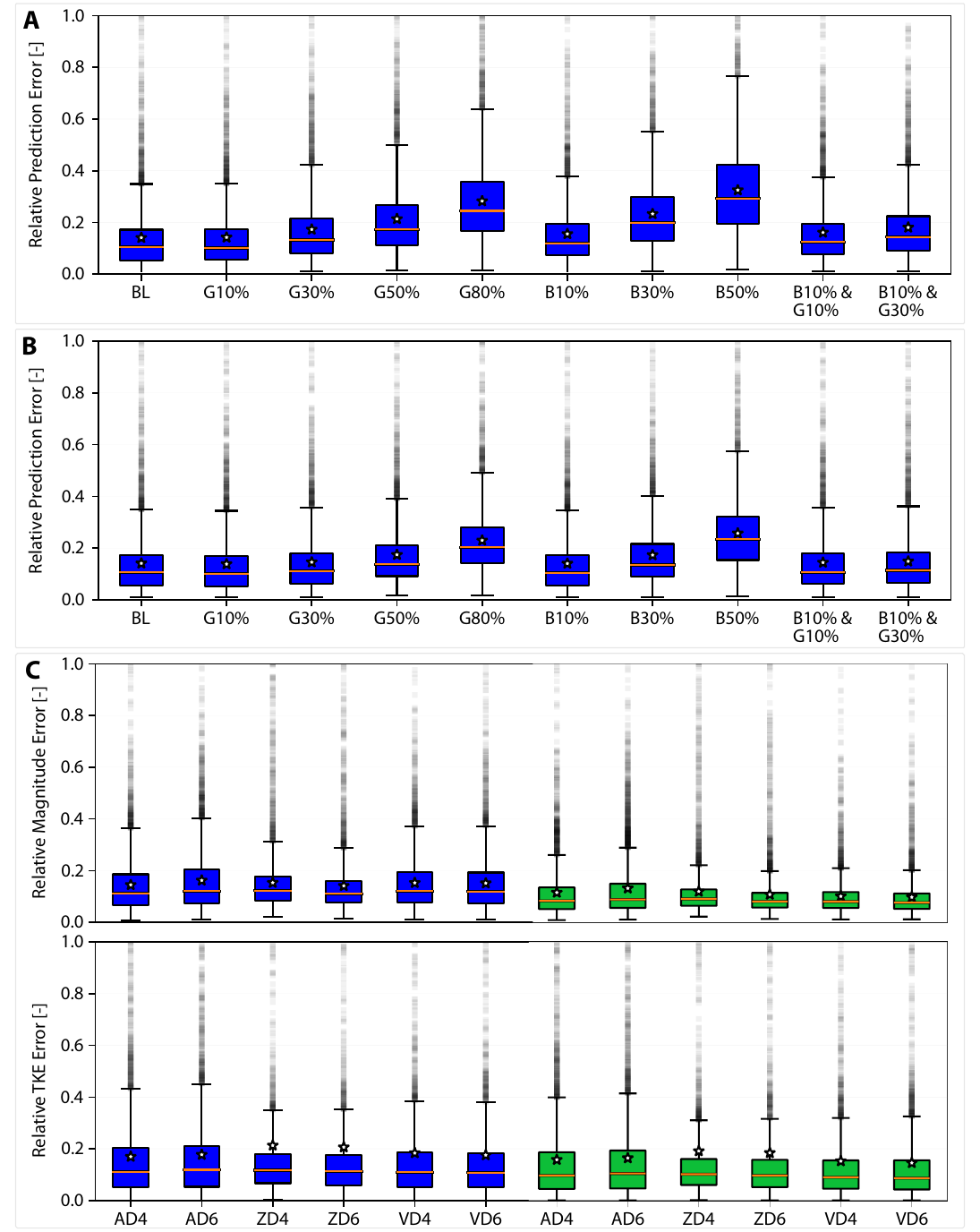}
    \caption[Noise experiment]{(\textbf{A}) Models trained with different levels of Gaussian noise and biases and evaluated with the same noise distribution used during training. (\textbf{B}) The same models as in \textbf{A} but evaluated without noise on the input data. (\textbf{C}) Wind magnitude and \ac{TKE} relative prediction errors of the \wcnn\ variants on the \ac{CFD} test set on the full domain (left, blue). In contrast to the velocity errors, excluding the closest cells to the terrain does not change the prediction error (right, green) for the \ac{TKE}.}
    \label{fig:noise_study}
\end{figure}

\paragraph{Input noise ablation study}
Realistic wind measurements are subject to noise. We model the sensor noise with a zero-mean Gaussian distribution and the sensor miscalibration with a constant bias. We evaluated the robustness of the model to different levels of such noisy input. Doing so we trained multiple models (BL architecture) with varying levels of input noise. We varied the standard deviation of the Gaussian noise between \qty{0}{\percent} and \qty{80}{\percent} of the average flow magnitude of the respective sample; we varied the bias between \qty{0}{\percent} and \qty{50}{\percent} of the flow magnitude. We then evaluated the models in two ways: First we evaluated them on the test set with the same noise distribution they observed during training. Since this is not a fair comparison, as predicting with high-noise levels is more difficult than low-noise data, we also evaluated all models against perfect data (no noise added). The results of the experiment are displayed in Fig.~\ref{fig:noise_study}~A). In general, higher input noise indicates higher prediction errors, but up to a level of \qty{10}{\percent} Gaussian noise and bias we observed similar errors. When evaluating the models on the perfect input we can see that the low-level noise models (up to \qty{10}{\percent} bias and \qty{30}{\percent} Gaussian noise) perform comparably to the baseline model trained without noise (Fig.~\ref{fig:noise_study}~B)). Thus training models with a too high noise level will also negatively impact the performance when they are provided with perfect input data.

\paragraph{CFD prediction results of different WindSeer variants}
In our evaluation we considered six variations of \wcnn\ [ZD4, ZD6, AD4, AD6, VD4, VD6] by varying the fill value and network depth. The fill indicator (Z, A, V) indicates how the wind speed input channels are filled for the cells with no measurements. We tested fill values of: zero (Z), the average of all measurements per channel (A), and the Voronoi tessellation presented in \cite{fukami2021global} (V) (essentially the nearest measurement value). The network depth indicates the number of pooling/upsampling layers in the encoder/decoder of \wcnn\ and we evaluated depths of four (D4) and six (D6) resulting in receptive field sizes of 175 and 703 respectively. The models were trained using the Adam optimizer \cite{kingma2015adam} for 3000 epochs except for AD6, where the model after 1000 epochs was chosen as further training showed increasing validation loss, suggesting over-fitting.

\begin{figure}
    \centering
    \includegraphics[width=0.6\textwidth]{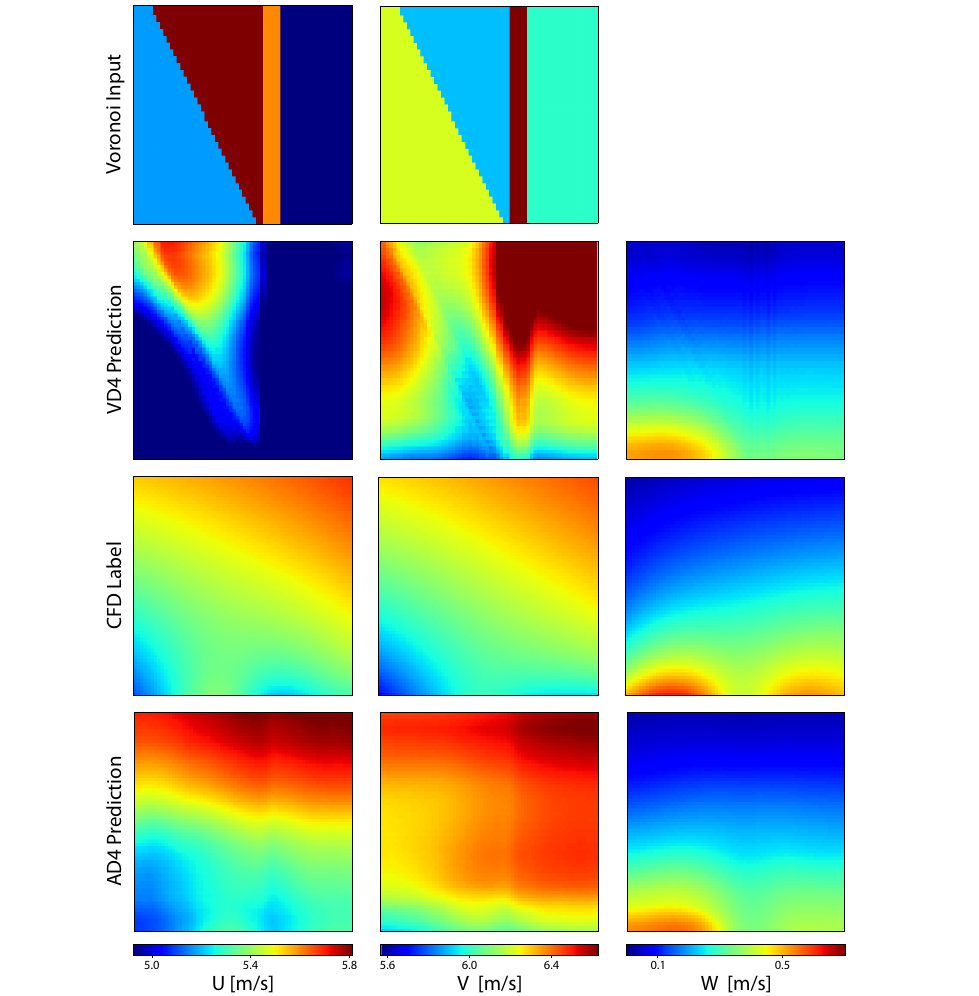}
    \caption[Voronoi Artifacts]{A slice through the domain showing the input to the Voronoi tessellation \wcnn\ variant (VD4) and the resulting prediction containing strong artifacts along the cell edges. The CFD label flow and the AD4 prediction. The AD4 prediction still shows some artifacts due to the input measurements albeit with much smaller significance.}
    \label{fig:voronoi_artifacts}
\end{figure}

We used the same input noise distribution as observed during training (Gaussian noise and random bias). Fig.~\ref{fig:noise_study} C) shows the distribution of the relative velocity magnitude and \ac{TKE} prediction errors over the full flow domain on the left side (blue) and excluding the lowest four cells above the terrain on the right side (green). These latter results (equivalent to only scoring the network output above an altitude of \qty{46}{m}) illustrate the predictive performance for realistic \ac{sUAV} flight regimes. There, all \wcnn\ variants produced more accurate wind velocity predictions (median error reduction AD4: \qty{11.1}{\percent} to \qty{8.3}{\percent}, AD6: \qty{12.0}{\percent} to \qty{8.9}{\percent}, ZD4: \qty{12.2}{\percent} to \qty{9.0}{\percent}, ZD6: \qty{11.0}{\percent} to \qty{8.1}{\percent}, VD4: \qty{12.1}{\percent} to \qty{8.0}{\percent}, VD6: \qty{11.9}{\percent} to \qty{7.6}{\percent}). In contrast to the velocity errors, the \ac{TKE} predictions do not significantly change on the reduced prediction volume since the computed \ac{TKE} values close to the terrain tend to be smoother than the velocity values, thus suffering less from resolution differences between \wcnn\ and the \ac{CFD} simulations. All the \wcnn\ variants result in a similar median between \qty{10.8}{\percent} to  \qty{11.9}{\percent}. Depending on the metric different models perform best. The averaging input models score the lowest mean error while the Voronoi variants yield the lowest median error. The zero-fill variants are most consistent with the lowest 75th percentile.

Overall, as evident in Fig.~\ref{fig:noise_study} C), there is no significant performance difference between the metrics of the \wcnn\ variants. However, a qualitative assessment of the predicted wind fields reveals that the Voronoi tessellation models (VD4 and VD6) show strong artifacts along the partition borders in certain cases. In contrast, the other \wcnn\ variants either do not exhibit such artifacts or are effected at a much smaller scale. These artifacts are a result of the noisy measurements being close to each other resulting in large differences between the Voronoi partitions. In 
Fig.~\ref{fig:voronoi_artifacts} we show one such case and the resulting predictions for the VD4 and AD4 \wcnn\ variants. These results indicate that while Voronoi tessellation has been shown to work with sparse input data for flow prediction~\cite{fukami2021global}, in our setting with highly noisy measurements from only a small sub-region of the domain, this input modality can result in predictions containing artifacts. Thus, we further evaluate the artifact-free \wcnn\ variants on the real wind data (AD4, AD6, ZD4, ZD6).

\paragraph{Measurement campaign results of different \wcnn{} variants}
We evaluated the performance of the different \wcnn{} variants on the data from the Bolund, Askervein, and \Perdigao{} data and report the prediction error and correlation averaged for certain wind cases, as in Tab.~\ref{tab:error_real_wind}. The changes in the prediction grid as outlined in Sec.~\ref{sec:method_measurement_campaigns} resulted in higher sparsity levels in the range of \qty{3.5e-6}{\percent} to \qty{3.2e-5}{\percent} compared to the training density of \qty{1.1e-3}{\percent} to \qty{0.19}{\percent}. The model variants using average-filling (AD4, AD6) could generalize to this much sparser input data in contrast to the zero-fill models (ZD4, ZD6), which severely underpredicted the wind regardless of the measurement values. Thus, we only compared the two performant \wcnn\ variants against an averaging baseline (AVG) that assumes the wind and \ac{TKE} are constant and predicts the average of all measurements over the full domain and report the prediction errors and correlations in Tab.~\ref{tab:app_error_meas_campaigns}.
The AD4 variant resulted in better wind magnitude predictions, while the AD6 predicted the vertical wind better. The \ac{TKE} is predicted with a lower error with the AD6 variant but also with lower correlation values compared to AD4. Overall, in most cases both \wcnn\ variants performed similarly to each other, explaining the small difference in the averaged error over all cases for the three metrics. We chose the AD4 variant as our \wcnn{} model as it did not show the over-fitting during training.

\begin{table}
    \centering
    \fontsize{6pt}{6pt}\selectfont
    {\tabulinesep=0.4mm
    \begin{tabu}{l l | c c c | c c c | c c c}
    \hline
     &  & \multicolumn{3}{c|}{Error S $[m/s]$} & \multicolumn{3}{c|}{Error W $[m/s]$} & \multicolumn{3}{c}{Error TKE $[m^2/s^2]$} \\
    Terrain & Case  & AVG & AD4 & AD6 & AVG & AD4 & AD6 & AVG & AD4 & AD6\\
    \hline
    \multirow{4}{*}{Bolund} & 90 & 1.80 & \textbf{1.58} & 1.61 & 0.85 & \textbf{0.58} & 0.62 & 1.91 & 1.33 & \textbf{1.19}\\
    & 239 & 2.80 & 2.50 & \textbf{2.49} & 0.66 & \textbf{0.34} & 0.37 & 2.67 & 1.68 & \textbf{1.66} \\ 
    & 255 & 3.24 & \textbf{2.47} & 2.61 & 0.85 & \textbf{0.44} & 0.46 & 3.43 & 2.12 & \textbf{2.06} \\ 
    & 270 & 3.77 & \textbf{2.79} & 2.94 & 0.95 & 0.51 & \textbf{0.48} & 5.14 & 3.43 & \textbf{3.35} \\
    \hline
    \multirow{13}{*}{Askervein} & TU25 & 2.58 & \textbf{2.39} & 2.44 & 1.10 & 0.37 & \textbf{0.31} & 0.61 & 0.41 & \textbf{0.34} \\ 
    & TU30A & 1.14 & \textbf{0.98} & 1.20 & 0.41 & 0.26 & \textbf{0.25} & 1.38 & 0.72 & \textbf{0.54} \\
    & TU30B & 1.80 & \textbf{1.46} & 2.13 & 0.51 & 0.41 & \textbf{0.39} & 2.82 & 1.40 & \textbf{1.18} \\ 
    & TU01A & 3.26 & \textbf{2.83} & 3.21 & 1.41 & 0.52 & \textbf{0.46} & 1.89 & 1.06 & \textbf{0.96} \\ 
    & TU01B & 3.24 & \textbf{2.74} & 3.12 & 1.37 & 0.48 & \textbf{0.42} & 1.64 & 0.98 & \textbf{0.87} \\ 
    & TU01C & 3.55 & \textbf{3.08} & 3.42 & 1.23 & 0.45 & \textbf{0.41} & 1.17 & 0.72 & \textbf{0.68} \\
    & TU01D & 4.21 & \textbf{3.71} & 4.04 & 1.26 & 0.47 & \textbf{0.42} & 1.62 & 1.17 & \textbf{0.99} \\
    & TU03A & 5.29 & \textbf{4.70} & 5.00 & 1.74 & 0.64 & \textbf{0.55} & 2.04 & 1.31 & \textbf{1.15} \\ 
    & TU03B & 4.90 & \textbf{4.41} & 4.70 & 1.54 & 0.54 & \textbf{0.46} & 1.82 & 1.21 & \textbf{1.04} \\ 
    & TU05A & 1.91 & \textbf{1.89} & 1.92 & 0.76 & 0.31 & \textbf{0.27} & 1.73 & 0.99 & \textbf{0.76} \\ 
    & TU05B & 1.18 & \textbf{1.00} & 1.12 & 0.31 & \textbf{0.26} & 0.28 & 1.40 & 0.63 & \textbf{0.51} \\
    & TU05C & 0.93 & \textbf{0.93} & 1.01 & 0.34 & 0.25 & \textbf{0.25} & 1.09 & 0.43 & \textbf{0.39} \\
    & TU07B & 3.42 & 3.27 & \textbf{3.18} & 1.59 & 0.49 & \textbf{0.41} & 2.44 & 1.85 & \textbf{1.43} \\
    \hline
    \multirow{4}{*}{\shortstack[l]{\Perdigao{}\\ 2017-05-09}} & 13:30-13:35 & 2.91 & 2.27 & \textbf{2.17} & 0.85 & \textbf{0.57} & 0.58 & - & - & - \\
    & 17:10:17:15 & 4.41 & 3.33 & \textbf{3.12} & 1.22 & 0.88 & \textbf{0.86} & - & - & - \\
    & 17:00-18:00 & 3.06 & 2.37 & \textbf{2.24} & 0.80 & \textbf{0.56} & 0.58 & - & - & - \\
    \hline
    \multirow{3}{*}{\shortstack[l]{\Perdigao{}\\ 2017-05-12}} & 01:00-02:00 & 2.82 & 2.31 & \textbf{2.23} & 0.52 & 0.42 & \textbf{0.40} & - & - \\ 
    & 17:00-18:00 & 2.76 & 2.23 & \textbf{2.11} & 0.70 & 0.57 & \textbf{0.47} & - & - & - \\
    & 19:45-19:50 & 1.15 & \textbf{0.90} & 0.91 & 0.21 & \textbf{0.16} & 0.18 & - & - & - \\
    \hline
    \multirow{4}{*}{\shortstack[l]{\Perdigao{}\\ 2017-05-16}} & 07:00-08:00 & 1.58 & 1.16 & \textbf{1.15} & 0.27 & 0.27 & \textbf{0.22} & - & - \\
    & 11:40-11:45 & 0.85 & 0.77 & \textbf{0.76} & 0.33 & 0.27 & \textbf{0.27} & - & - & - \\
    & 12:40-12:45 & 0.86 & 0.75 & \textbf{0.74} & 0.29 & 0.22 & \textbf{0.22} & - & - & - \\
    & 20:00-21:00 & 1.35 & \textbf{0.97} & 1.04 & \textbf{0.22} & 0.31 & 0.29 & - & - & - \\
    \hline
    \multirow{3}{*}{\shortstack[l]{\Perdigao{}\\ 2017-05-18}} & 14:35-14:40 & 2.14 & 1.82 & \textbf{1.75} & 0.41 & 0.36 & \textbf{0.33} & - & - & - \\
    & 20:00-21:00 & 1.54 & \textbf{1.08} & 1.11 & \textbf{0.17} & 0.19 & 0.20 & - & - & - \\
    & 22:00-23:00 & 1.52 & \textbf{1.13} & 1.18 & 0.17 & 0.17 & \textbf{0.16} & - & - & - \\
    \hline
    \multirow{3}{*}{\shortstack[l]{\Perdigao{}\\ 2017-05-20}} & 03:15-03:20 & 3.59 & \textbf{2.63} & 2.77 & \textbf{0.41} & 0.55 & 0.46 & - & - & - \\
    & 10:00-11:00 & 2.20 & 1.84 & \textbf{1.75} & 0.51 & 0.42 & \textbf{0.38} & - & - & - \\
    & 12:20-12:25 & 1.93 & 1.71 & \textbf{1.61} & 0.58 & 0.43 & \textbf{0.41} & - & - & - \\
    \hline
    \multirow{3}{*}{\shortstack[l]{\Perdigao{}\\ 2017-06-08}} & 00:00-01:00 & 2.68 & \textbf{1.87} & 2.03 & 0.35 & 0.38 & \textbf{0.33} & - & - & - \\
    & 12:40-12:45 & 1.20 & 0.90 & \textbf{0.87} & 0.31 & \textbf{0.22} & 0.24 & - & - & - \\
    & 14:00-15:00 & 2.49 & 1.77 & \textbf{1.64} & 0.74 & \textbf{0.42} & 0.43 & - & - & - \\
    \hline
    \multicolumn{2}{l|}{Total average} & 2.50 & \textbf{2.07} & 2.13 & 0.72 & 0.41 & \textbf{0.38} & 2.05 & 1.26 & \textbf{1.12} \\
    \hline
    \hline
     &  & \multicolumn{3}{c|}{Correlation S} & \multicolumn{3}{c|}{Correlation W} & \multicolumn{3}{c}{Correlation TKE} \\
    Terrain & Case & AVG & AD4 & AD6 & AVG & AD4 & AD6 & AVG & AD4 & AD6\\
    \hline
    \multirow{4}{*}{Bolund} & 90 & - & \textbf{0.72} & 0.67 & - & 0.50 & \textbf{0.64} & - & 0.58 & \textbf{0.60} \\ 
    & 239 & - & 0.68 & \textbf{0.73} & - & \textbf{0.76} & 0.75 & - & 0.86 & \textbf{0.93} \\ 
    & 255 & - & 0.82 & \textbf{0.85} & - & 0.72 & \textbf{0.76} & - & 0.82 & \textbf{0.89} \\ 
    & 270 & - & 0.85 & \textbf{0.91} & - & 0.78 & \textbf{0.84} & - & 0.73 & \textbf{0.78} \\
    \hline
    \multirow{13}{*}{Askervein} & TU25 & - & \textbf{0.65} & 0.60 & - & 0.90 & \textbf{0.96} & - & \textbf{0.89} & 0.88 \\ 
    & TU30A & - & 0.62 & \textbf{0.68} & - & 0.58 & \textbf{0.74} & - & 0.42 & \textbf{0.54} \\
    & TU30B & - & \textbf{0.73} & 0.71 & - & \textbf{0.64} & 0.63 & - & 0.23 & \textbf{0.50} \\ 
    & TU01A & - & \textbf{0.77} & 0.52 & - & 0.91 & \textbf{0.92} & - & \textbf{0.85} & 0.46 \\ 
    & TU01B & - & \textbf{0.79} & 0.53 & - & 0.92 & \textbf{0.93} & - & \textbf{0.87} & 0.53 \\
    & TU01C & - & \textbf{0.78} & 0.46 & - & 0.92 & \textbf{0.93} & - & \textbf{0.90} & 0.64 \\
    & TU01D & - & \textbf{0.79} & 0.54 & - & 0.93 & \textbf{0.94} & - & \textbf{0.93} & 0.81 \\
    & TU03A & - & \textbf{0.78} & 0.65 & - & 0.93 & \textbf{0.95} & - & \textbf{0.98} & 0.94 \\ 
    & TU03B & - & \textbf{0.77} & 0.61 & - & 0.92 & \textbf{0.95} & - & \textbf{0.90} & 0.89 \\ 
    & TU05A & - & \textbf{0.62} & 0.53 & - & 0.89 & \textbf{0.91} & - & \textbf{0.40} & 0.31 \\ 
    & TU05B & - & \textbf{0.79} & 0.69 & - & 0.48 & \textbf{0.53} & - & \textbf{0.04} & -0.09 \\
    & TU05C & - & \textbf{0.66} & 0.53 & - & 0.58 & \textbf{0.60} & - & \textbf{0.14} & -0.05 \\
    & TU07B & - & \textbf{0.70} & 0.66 & - & 0.90 & \textbf{0.97} & - & 0.40 & \textbf{0.41} \\
    \hline
    \multirow{4}{*}{\shortstack[l]{\Perdigao{}\\ 2017-05-09}} & 13:32:30 & - & \textbf{0.82} & 0.82 & - & 0.53 & \textbf{0.57} & - & - & -\\
    & 17:12:30 & - & 0.48 &\textbf{ 0.80} & - & 0.35 & \textbf{0.48} & - & - & - \\
    & 17:00-18:00 & - & 0.77 & \textbf{0.77} & - & 0.50 & \textbf{0.54} & - & - & - \\
    \hline
    \multirow{3}{*}{\shortstack[l]{\Perdigao{}\\ 2017-05-12}} & 01:00-02:00 & - & \textbf{0.76} & 0.74 & - & 0.45 & \textbf{0.51} & - & - & - \\ 
    & 17:00-18:00 & - & \textbf{0.81} & 0.81 & - & 0.57 & \textbf{0.61} & - & - & - \\
    & 19:45-19:50 & - & 0.71 & \textbf{0.72} & - & \textbf{0.57} & 0.53 & - & - & - \\
    \hline
    \multirow{4}{*}{\shortstack[l]{\Perdigao{}\\ 2017-05-16}} & 07:00-08:00 & - & 0.65 & \textbf{0.68} & - & \textbf{0.67} & 0.63 & - & - & - \\
    & 11:40-11:45 & - & 0.51 & \textbf{0.55} & - & \textbf{0.22} & 0.18 & - & - & - \\
    & 12:40-12:45 & - & \textbf{0.48} & 0.45 & - & 0.30 & \textbf{0.31} & - & - & - \\
    & 20:00-21:00 & - & \textbf{0.68} & 0.64 & - & 0.21 & \textbf{0.25} & - & - & - \\
    \hline
    \multirow{3}{*}{\shortstack[l]{\Perdigao{}\\ 2017-05-18}} & 14:35-14:40 & - & 0.70 & \textbf{0.72} & - & 0.33 & \textbf{0.39} & - & - & - \\
    & 20:00-21:00 & - & \textbf{0.84} & 0.80 & - & 0.12 & \textbf{0.19} & - & - & - \\
    & 22:00-23:00 & - & \textbf{0.74} & 0.69 & - & 0.42 & \textbf{0.50} & - & - & - \\
    \hline
    \multirow{3}{*}{\shortstack[l]{\Perdigao{}\\ 2017-05-20}} & 03:15-03:20 & - & 0.61 & \textbf{0.64} & - & \textbf{0.30} & 0.30 & - & - & - \\
    & 10:00-11:00 & - & 0.71 & \textbf{0.76} & - & \textbf{0.46} & 0.44 & - & - & - \\
    & 12:20-12:25 & - & 0.66 & \textbf{0.71} & - & \textbf{0.45} & 0.42 & - & - & - \\
    \hline
    \multirow{3}{*}{\shortstack[l]{\Perdigao{}\\ 2017-06-08}} & 00:00-01:00 & - & \textbf{0.69} & 0.67 & - & \textbf{0.44} & 0.42 & - & - & - \\
    & 12:40-12:45 & - & 0.77 & \textbf{0.77} & - & 0.46 & \textbf{0.46} & - & - & - \\
    & 14:00-15:00 & - & \textbf{0.82} & 0.82 & - & 0.58 & \textbf{0.62} & - & - & - \\
    \hline
    \multicolumn{2}{l|}{Total average} & - & \textbf{0.71} & 0.68 & - & 0.59 & \textbf{0.62} & - & \textbf{0.64} & 0.59 \\
    \hline
    \end{tabu}
    \caption[Absolute prediction errors.]{\textbf{Measurement campaigns error results}. Absolute prediction errors and correlations for the velocity magnitude (S), vertical wind component (W), and turbulence kinetic energy (TKE) on the measurement campaign datasets of the AD4 and AD6 models compared to the averaging baseline (AVG).}}
    \label{tab:app_error_meas_campaigns}
\end{table}

\clearpage
\paragraph{\ac{sUAV} results of different \wcnn{} variants}
We present the error metrics for all flights and the different model variants in Tab.~\ref{tab:flight_static_loiter}. All models consistently struggle at predicting the horizontal wind while the vertical wind prediction is much more accurate compared to the baseline. The average-filling models strongly rely on the averaged measurement, thus providing a good representation of the overall flow state. However, in the flight experiments with complex topography, the measured wind does not have this property, therefore the zero-fill models (ZD4, ZD6), that learnt to better encode the measurement locations, outperform the average-filling (AD4, AD6) variants. In this set of experiment we see a slight trend that increasing network depth seems to improve the prediction quality.

\begin{table}
    \centering
    {\tabulinesep=0.4mm
    \begin{tabu}{l l | c c c | c c c}
    \hline
     &  & \multicolumn{3}{c|}{Mean Absolute Error} & \multicolumn{3}{c}{Correlation} \\
    Flight & Model  & $W_{hor}$ [m/s] & $\Psi_{hor}$  [$\deg$] & $W_z$  [m/s] & $W_{hor}$ & $\Psi_{hor}$ & $W_z$\\
    \hline
    \multirow{5}{*}{Chasseral 1} & AVG & 0.62 & \textbf{7.22} & 0.53 & - & - & - \\ 
    & AD4 & 0.77 & 7.84 & 0.66 & 0.54 & \textbf{-0.09} & 0.94 \\ 
    & AD6 & \textbf{0.61} & 7.49 & 0.55 & \textbf{0.77} & -0.17 & \textbf{0.95} \\ 
    & ZD4 & 0.87 & 9.18 & 0.41 & 0.26 & -0.13 & 0.95 \\ 
    & ZD6 & 0.84 & 9.19 & \textbf{0.38} & 0.33 & -0.37 & 0.95 \\
    \hline
    \multirow{5}{*}{Chasseral 2} & AVG & 0.57 & \textbf{13.6} & 0.50 & - & - & - \\ 
    & AD4 & 0.66 & 14.7 & 0.48 & \textbf{0.50} & \textbf{-0.35} & \textbf{0.90} \\ 
    & AD6 & \textbf{0.48} & 14.1 & 0.47 & 0.48 & -0.46 & 0.87 \\ 
    & ZD4 & 0.73 & 17.2 & 0.35 & 0.49 & -0.42 & 0.78 \\ 
    & ZD6 & 0.74 & 16.8 & \textbf{0.31} & 0.49 & -0.60 & 0.80 \\
    \hline
    \multirow{5}{*}{Chasseral 3} & AVG & 0.55 & \textbf{9.1} & 0.49 & - & - & - \\ 
    & AD4 & 0.56 & 9.8 & 0.39 & 0.43 & \textbf{-0.21} & 0.82 \\ 
    & AD6 & \textbf{0.55} & 9.5 & 0.40 & \textbf{0.51} & -0.33 & 0.84 \\ 
    & ZD4 & 0.67 & 14.0 & 0.32 & 0.32 & -0.43 & 0.80 \\ 
    & ZD6 & 0.65 & 12.4 & \textbf{0.30} & 0.37 & -0.38 & \textbf{0.84} \\
    \hline
    \multirow{5}{*}{Oberalppass} & AVG & \textbf{0.55} & 7.0 & 0.55 & - & - & - \\ 
    & AD4 & 1.05 & 19.7 & 0.30 & 0.28 & -0.86 & \textbf{0.81} \\ 
    & AD6 & 3.12 & 7.5 & 0.35 & \textbf{0.43} & \textbf{0.85} & 0.76 \\ 
    & ZD4 & 0.65 & \textbf{5.8} & \textbf{0.33} & -0.24 & 0.62 & 0.78 \\ 
    & ZD6 & 0.77 & 7.9 & 0.34 & -0.46 & -0.91 & 0.77 \\
    \hline
    \multirow{5}{*}{Gotthardpass} & AVG & \textbf{1.00} & \textbf{7.5} & 0.21 & - & - & - \\ 
    & AD4 & 2.55 & 64.3 & 0.57 & 0.26 & 0.75 & 0.21 \\ 
    & AD6 & 1.41 & 58.3 & 1.06 & \textbf{0.71} & \textbf{0.98} & 0.48 \\ 
    & ZD4 & 1.40 & 7.9 & 0.20 & -0.06 & 0.97 & 0.48 \\ 
    & ZD6 & 1.19 & 7.8 & \textbf{0.17} & 0.33 & 0.63 & \textbf{0.88} \\
    \hline
    \hline
    \multirow{5}{*}{\shortstack{All Chasseral \\ Flights}} & AVG & 0.58 & \textbf{9.98} & 0.51 & - & - & - \\ 
    & AD4 & 0.66 & 10.78 & 0.51 & 0.49 & -0.22 & \textbf{0.89} \\ 
    & AD6 & \textbf{0.55} & 10.36 & 0.47 & \textbf{0.55} & -0.32 & \textbf{0.89} \\ 
    & ZD4 & 0.76 & 13.46 & 0.36 & 0.36 & -0.33 & 0.84 \\ 
    & ZD6 & 0.74 & 10.60 & \textbf{0.34} & 0.19 & -0.11 & 0.86 \\
    \hline
    \multirow{5}{*}{All Flights} & AVG & \textbf{0.66} & \textbf{8.88} & 0.46 & - & - & - \\ 
    & AD4 & 1.12 & 23.27 & 0.48 & 0.40 & -0.15 & 0.74 \\ 
    & AD6 & 1.23 & 19.38 & 0.57 & \textbf{0.58} & \textbf{0.17} & 0.78 \\ 
    & ZD4 & 0.86 & 10.82 & 0.32 & 0.15 & 0.12 & 0.76 \\ 
    & ZD6 & 0.84 & 9.5 & \textbf{0.30} & 0.09 & -0.13 & \textbf{0.85} \\
    \end{tabu}}
    \caption[correlation]{\textbf{\ac{sUAV} flight results}. The mean absolute error and the correlation for the horizontal wind magnitude $W_{hor}$, wind direction $\Psi_{hor}$, and vertical wind $W_z$ on the loiter-averaged data. The results are the average over all loiters for all planes for the respective flight.}
    \label{tab:flight_static_loiter}
\end{table}

\clearpage

\section{\ac{sUAV} airflow sensing}
\label{sec:app_airflow_sensing}
In this section we outline the design the airflow vanes used to estimate the full 3D wind aboard the \acp{sUAV}. We explain and evaluate the calibration procedure to account for the constant mounting offset and the dynamic aerodynamic biases.

\paragraph{Design and calibration of \ac{sUAV} airflow vanes}

\begin{figure}
    \centering
    \includegraphics[width=0.9\textwidth]{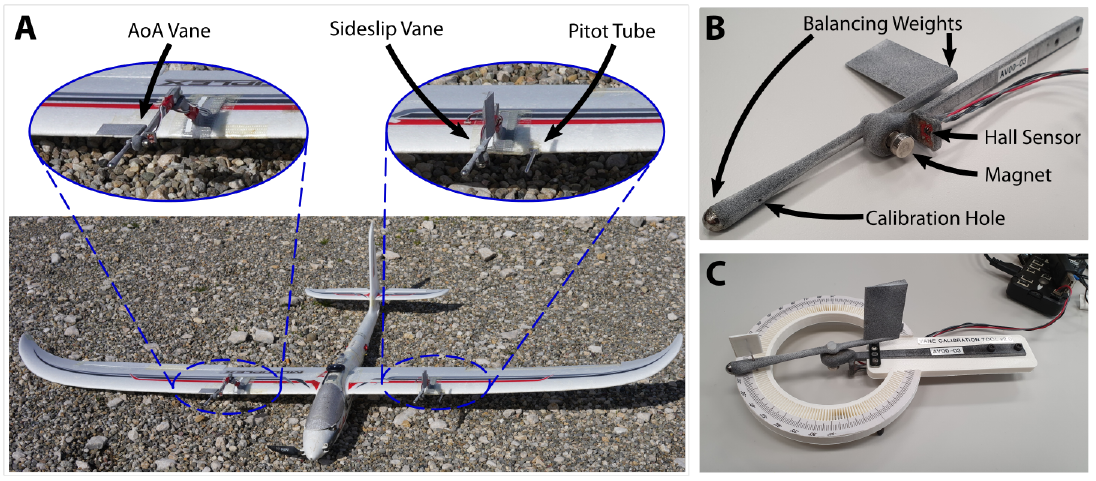}
    \caption[Airflow Sensing]{(\textbf{A}) The arrangement of the airflow sensors on the \ac{sUAV}. (\textbf{B}) Components of the airflow vanes. (\textbf{C}) Calibration tool used to determine the mapping from the magnetic flux density to the angle.}
    \label{fig:airflow_sensing}
\end{figure}
Two custom-designed wind vanes together with a pitot tube measure the 3D airflow. One vane measures the \ac{AoA} and the other one the \ac{AoS} of the airspeed vector relative to the \ac{sUAV} body reference frame. During flight the wings flex due to maneuvers or wind gusts causing measurement error on the vanes that are larger if the vanes are mounted further towards the wingtips. However, the prop wash makes any placement too close to the fuselage invalid since the vanes need to measure the undisturbed free flow. Therefore we place the vanes approximately one quarter of the wing length away from the fuselage (Fig.~\ref{fig:airflow_sensing}~A)).

The vane is 3D printed and balanced using metal weights at the front and back (Fig.~\ref{fig:airflow_sensing}~B)). Small ball bearings at the connection axis ensure little friction in the setup and fast response time to changing wind. A diametric radial magnet is mounted at the end of the connection axis resulting in a changing magnetic field (for different angles) that the Hall sensor measures.

\begin{figure} [b]
    \centering
    \includegraphics[width=0.8\textwidth]{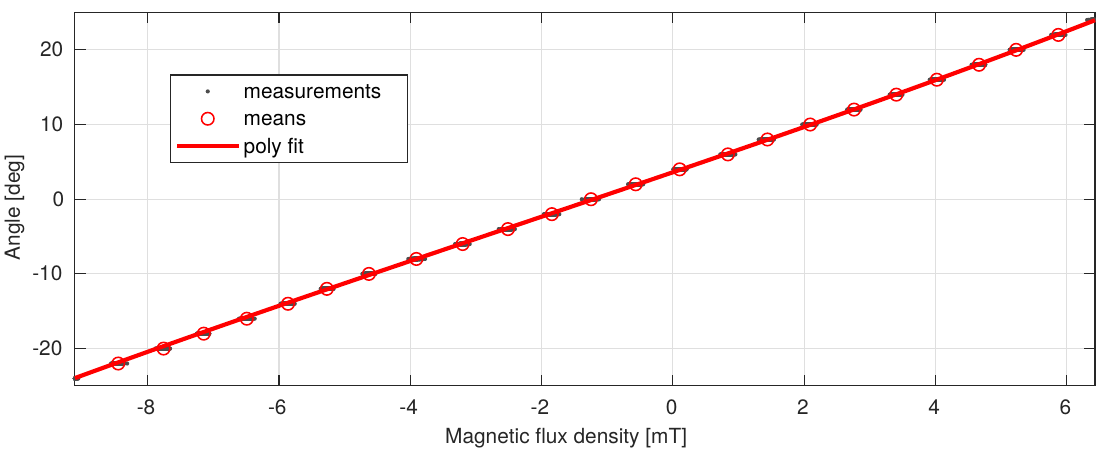}
    \caption[Hall Mapping]{Magnetic flux density to angle mapping for one wind vane with the measured data during the calibration procedure.}
    \label{fig:hall_mapping}
\end{figure}

The calibration tool, shown in Fig.~\ref{fig:airflow_sensing}~C), allowed us to accurately set the vanes to angles with \qty{2}{\degree} increments, thus gathering accurate data to determine the mapping from the magnetic flux density $B$. We calibrated each sensor for angles ranging from \qty{-24}{\degree} to \qty{24}{\degree} using a third-order polynomial function. Fig.~\ref{fig:hall_mapping} shows the measurements and the resulting fit for one wind vane.

\paragraph{\ac{sUAV} airflow sensing calibration}
Raw \ac{AoA} and \ac{AoS} measurements are subject to mounting errors as well as aerodynamic effects from the fuselage and the wing. We defined calibration functions based on wind tunnel data provided by Heinrich et al.~\cite{heinrich2021deepstall} to estimate the true airflow angles based on the sensor measurements $(\alpha_{raw}, \beta_{raw})$ in the relevant range (\ac{AoA} between $0^\circ$ to $15^\circ$, \ac{AoS} between $-10^\circ$ to $10^\circ$):
\begin{align}
    \alpha_{off} &= p_{\alpha,0} + p_{\alpha,1} \cdot \alpha_{raw} + p_{\alpha,2} \left(\alpha_{raw} + p_{\alpha,3} \right) \left(v_{Aspd} + p_{\alpha,4}  \right), \label{eq:alpha_cal}\\
    \beta_{off} &= p_{\beta,0} + p_{\beta,1} \left(v_{Aspd} + p_{\beta,2}  \right) \left(1 + \tanh \left(p_{\beta,3} \left(\alpha_{raw} + p_{\beta,4}\right)\right)\right) + \nonumber\\ 
    & p_{\beta,5} \cdot \beta_{raw} + p_{\beta,6} \cdot \tanh \left(p_{\beta,7} \left(\Phi + p_{\beta,8} \right) \right),\label{eq:beta_cal}
\end{align}
where the $p$ variables are free parameters. For the wind tunnel validation, the parameters were estimated by minimizing the \ac{MSE} between the sensor measurements and ground truth airflow angles (orientation of the aircraft using a tunnel-mounted sting, assumed to have very low angular position error). Fitting the wind tunnel data, the base functions result in \iac{MSE} for the \ac{AoA} of $0.45^\circ$ and $0.83^\circ$ for the \ac{AoS}.

\begin{figure}
    \centering
    \includegraphics[width=0.9\textwidth]{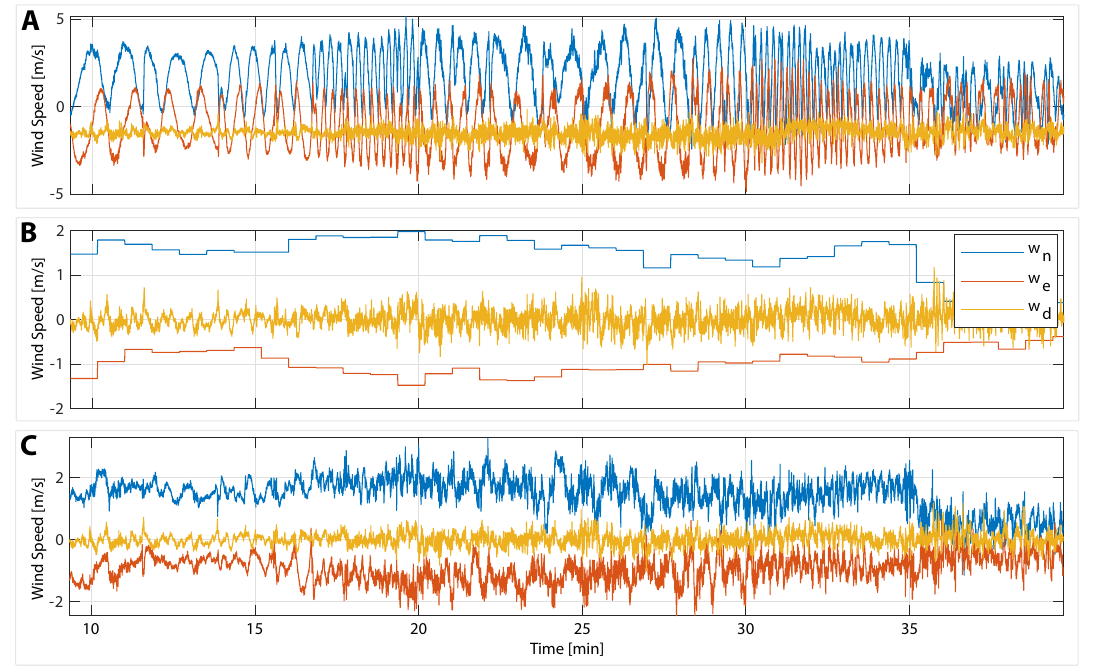}
    \caption[Airflow Calibration]{(\textbf{A}) The wind estimates based on the raw uncalibrated airflow sensor data. (\textbf{B}) The piecewise horizontal wind and zero-mean vertical wind as optimized during the calibration procedure. (\textbf{C}) The wind estimates after calibrating the airflow sensors.}
    \label{fig:airflow_calibration}
\end{figure}

However, due to variations in mounts and aircraft, this calibration could not be performed for every sensor installation. Thus, we further defined a calibration routine to estimate the parameters of Eq.~\ref{eq:alpha_cal}, \ref{eq:beta_cal} based on data gathered during a calibration flight, removing the need to calibrate every \ac{sUAV} with wind tunnel data. The underlying assumptions that ensure the parameters are observable are that the horizontal wind is piecewise constant and that there is no vertical wind during the calibration flight (calibration flights were performed in as calm flight conditions as possible, usually early morning). We also assume the estimated attitude and global position/velocity are accurate. To cover the different flight regimes our calibration flight consisted of counter-clockwise and clockwise loiter circles of different radii ranging from \qtyrange{30}{100}{\m} flown at different airspeeds (\qty{10}{\meter\per\second} to \qty{16}{\meter\per\second}). We then solved for the calibration parameters and the wind ($W_x$, $W_y$) by minimizing the error using a nonlinear least-squares solver in the wind triangle over the full flight:
\begin{equation}
    \boldsymbol{e} = R\left(\Phi, \Theta, \Psi\right)
    \begin{bmatrix}
           v_{Aspd} \\
           v_{Aspd} \cdot \tanh \left(\beta - \beta_{off}\right) + l_{x,\beta} \cdot \omega_z - l_{z,\beta} \cdot \omega_x\\
           v_{Aspd} \cdot \tanh \left(\alpha - \alpha_{off}\right) - l_{x,\alpha} \cdot \omega_y + l_{y,\alpha} \cdot \omega_x
     \end{bmatrix} +
     \begin{bmatrix}
           W_x \\
           W_y\\
           0
     \end{bmatrix}
     - \boldsymbol{v_{Gnd}},
\end{equation}
where $R\left(\Theta, \Phi, \Psi\right)$ is the rotation matrix based on the current attitude, $\boldsymbol{v_{Gnd}}$ the estimated ground speed vector, and $\omega_{(.)}$ the rotational speed around the respective axis. The offset from the vanes to the autopilot origin is denoted by $l_{x,\beta}$, $l_{z,\beta}$, $l_{x,\alpha}$, and $l_{y,\alpha}$.

Using the uncalibrated measurements from the airflow sensors results in strong oscillations of the estimated horizontal wind (strongly correlated to the loiter frequency) and a vertical estimate with a non-zero mean as visible in Fig.~\ref{fig:airflow_calibration}~A) as the sensors are located within the disturbed flow from the wing and airframe. The piecewise linear horizontal and zero-mean vertical wind fit as a result of the airflow calibration pipeline are shown Fig.~\ref{fig:airflow_calibration}~B). Although there is no constraint on the difference between the segments in the horizontal wind, the changes are relatively small. This stable, near-constant wind (magnitude and direction) reflects the forecast and observations made from the ground during the flight. The calibration reduces the estimated oscillations in the wind significantly and results in accurately measuring the zero-mean vertical wind (Fig.~\ref{fig:airflow_calibration}~C)). However, some correlation between the wind estimates and the loiter frequency remain, indicating that the calibration function could still be improved.

\begin{figure}
    \centering
    \includegraphics[width=1.0\textwidth]{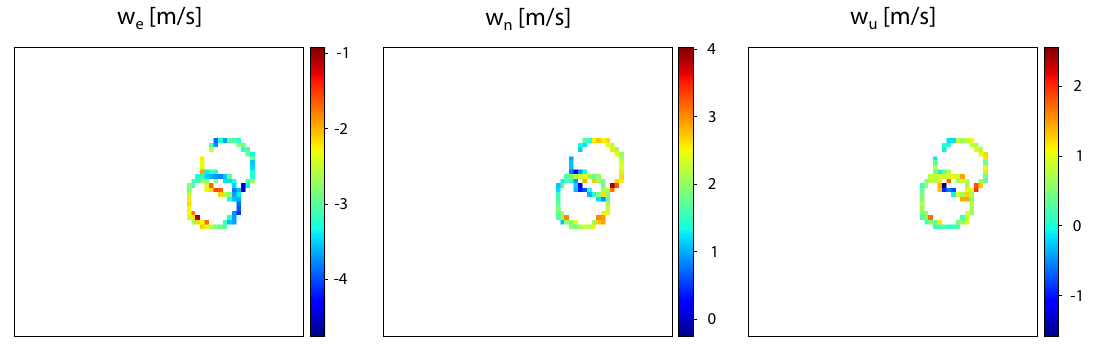}
    \caption[Airflow Calibration Eval]{A top down view of the binned wind measurements for one \ac{sUAV} for the first Chasseral flight for two loiters flown at the same altitude.}
    \label{fig:airflow_calibration_eval}
\end{figure}

\paragraph{Calibration quality}
In contrast to the calibration flights, in the actual data collection flights, the wind estimates from the \acp{sUAV} again show some oscillations strongly correlating with the loiter patterns in flight. However, for these data collection flights we expect the wind to vary across different locations so this could be correctly observed changes in the wind field. In Fig.~\ref{fig:airflow_calibration_eval} we display the binned wind observations from two loiters patterns flown at the same altitude next to each other. Especially for the horizontal wind measurements $w_e$ and $w_n$ we can see the same pattern repeating for both loiters with an amplitude of about \qty{1}{\meter\per\second} in each direction. This pattern indicates that we would expect errors in the horizontal measurements of about $\pm$\qty{1}{\meter\per\second}, which are comparable to the observed variation of the measurements between the \ac{sUAV} and within a single flight. For the vertical wind we do not see such repeating patterns, thus we expect a higher quality of these measurements.

The altitude for the calibration flight of \qty{540}{\meter} above mean sea level compared to altitudes of the data collection flights (\qtyrange{1600}{2200}{\meter}) results in \qtyrange{10.5}{15.1}{\%} lower air densities at the higher altitudes. Previous work has shown that density changes result in changing flow fields~\cite{ejeh2020airflowdensity, sayed2015airflowdensity}. This could, in part, explain the difficulty to accurately calibrating the airflow sensing if they are located within the disturbed flow field of the air-frame. Therefore, for future flights, the sensors should be placed further away from the wings and fuselage to minimize the aerodynamic disturbances on the sensors, and calibration flights performed at the same altitude as test flights.

\end{appendices}

\clearpage
\renewcommand\thefigure{\arabic{figure}}   

\begin{figure} [p]
    \centering
    \includegraphics[width=1.0\textwidth]{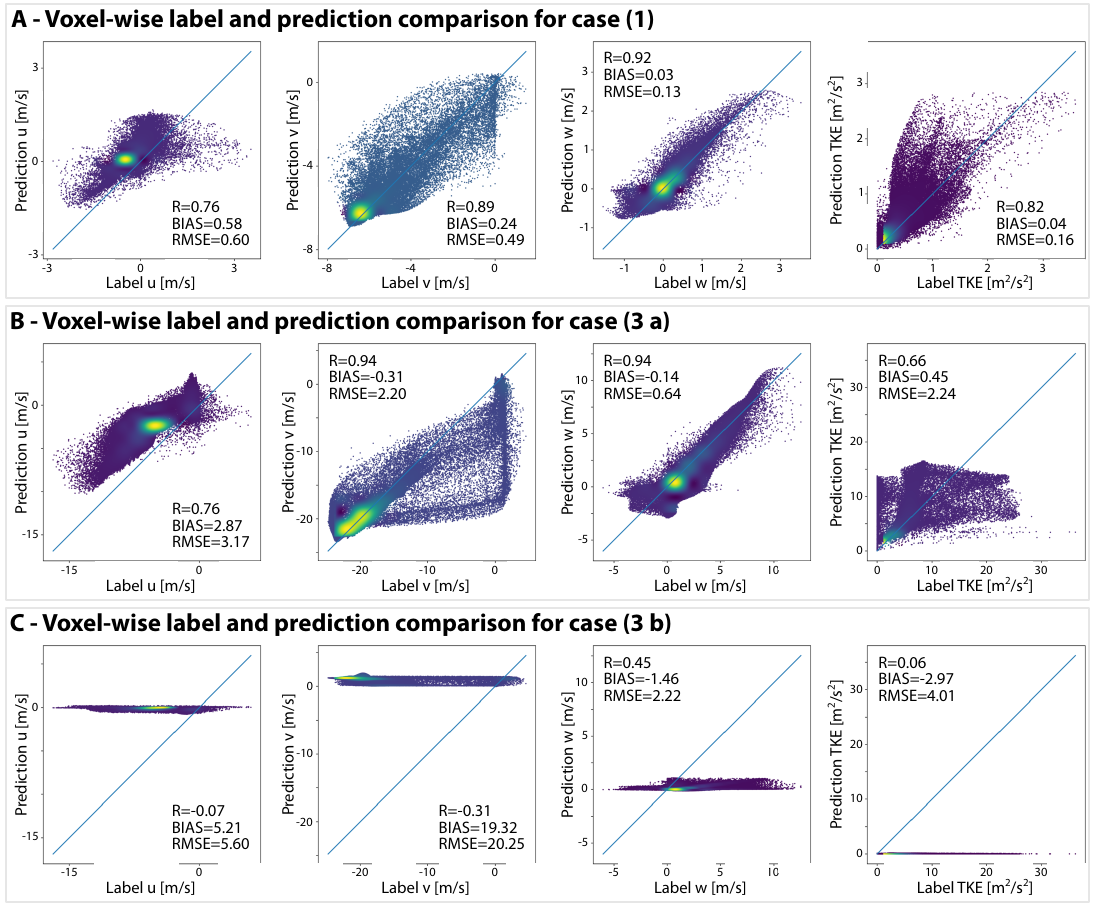}
    \caption[Appendix Scatter Plots]{\textbf{CFD experiments results}. Labels compared to the prediction for each cell and channel for the terrain and input measurements presented in \ref{fig:cfd_results} A (1) and (3).}
    \label{fig:app_scatter_plots}
\end{figure}

\begin{figure} [p]
    \centering
    \includegraphics[width=1.0\textwidth]{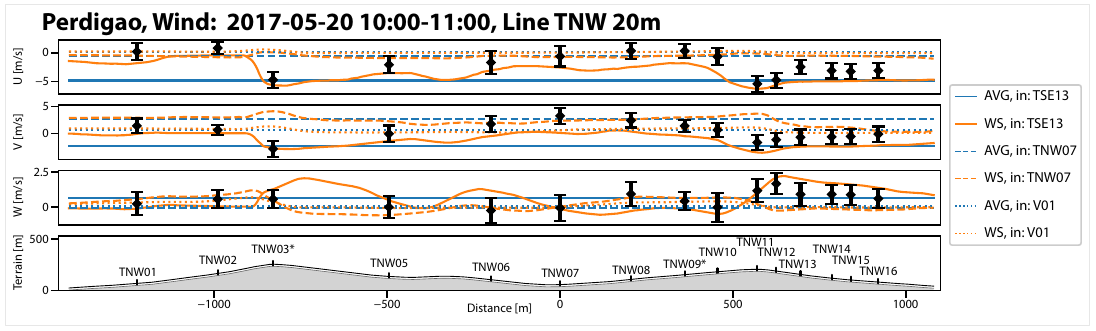}
    \caption[Results Measurement Campaigns]{\textbf{Measurement campaigns results}. Predictions and measurements along characteristic lines with a constant height for each experiment with the baseline averaging method (AVG) and \wcnn{} (WS). Three predictions using different input masts are shown. The asterisk * indicates that no measurement was available for that respective mast at the queried height and the closest one was picked. The uncertainty of the measurements is displayed by the standard deviation of the raw high-rate data. The measurements indicate a rotor between the two ridges, a flow pattern not present in the training data. Thus, \wcnn\ struggles to accurately representing the flow for the towers TNW06-TNW10.}
    \label{fig:meas_campaign_result_extended_data}
\end{figure}

\begin{figure} [p]
    \centering
    \includegraphics[width=1.0\textwidth]{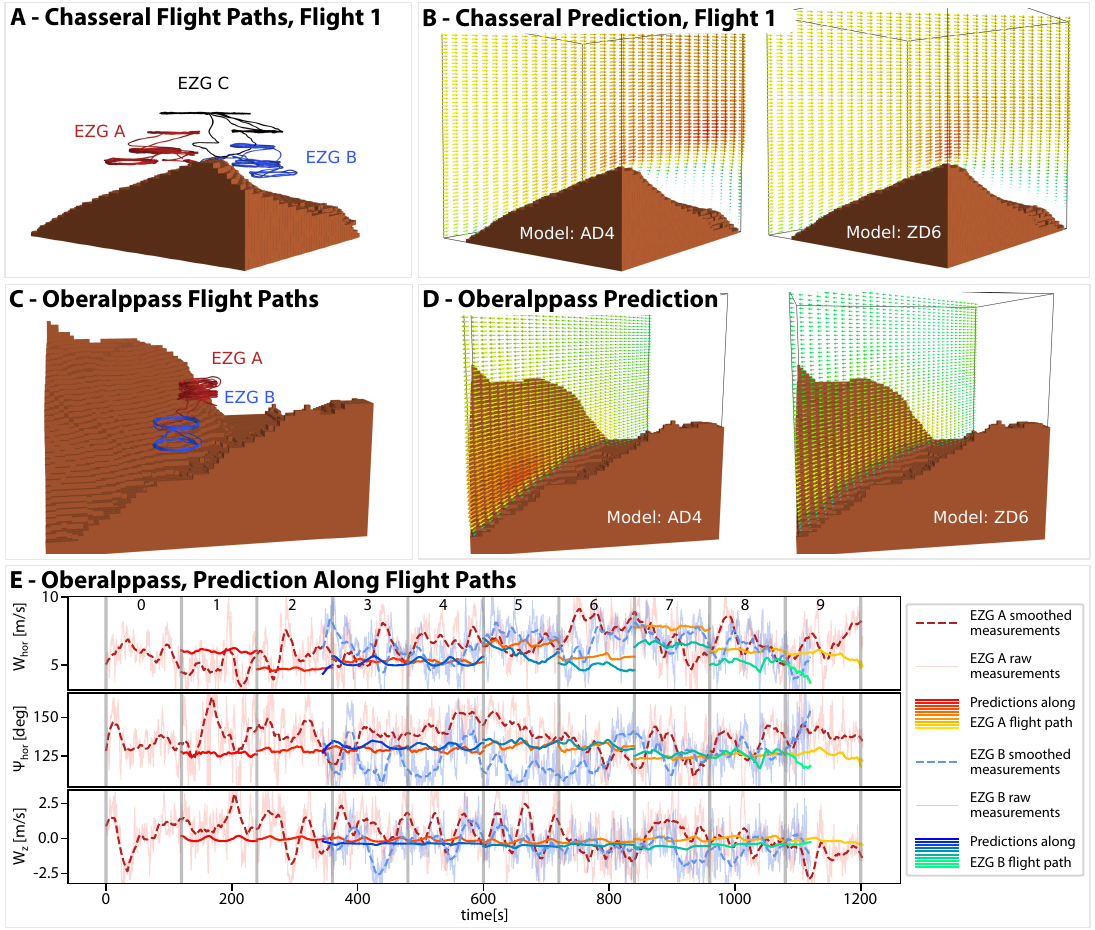}
    \caption[Flight Tests]{\textbf{\Ac{sUAV} flight experiment}. Prediction results and flight paths for two flight tests: Chasseral (\textbf{A}-\textbf{C}) and Oberalppass (\textbf{D}-\textbf{F}). The predictions along a slice are shown for the AD4 and ZD6 models (\textbf{B}, \textbf{E}). (\textbf{C}) and (\textbf{F}) show the sliding window predictions of ZD6 along the flight paths using the data from EZG A as input. Every \qty{120}{s} a prediction is made using the wind data from the previous window.}
    \label{fig:flight_tests}
\end{figure}

\begin{figure} [p]
    \centering
    \includegraphics[width=1.0\textwidth]{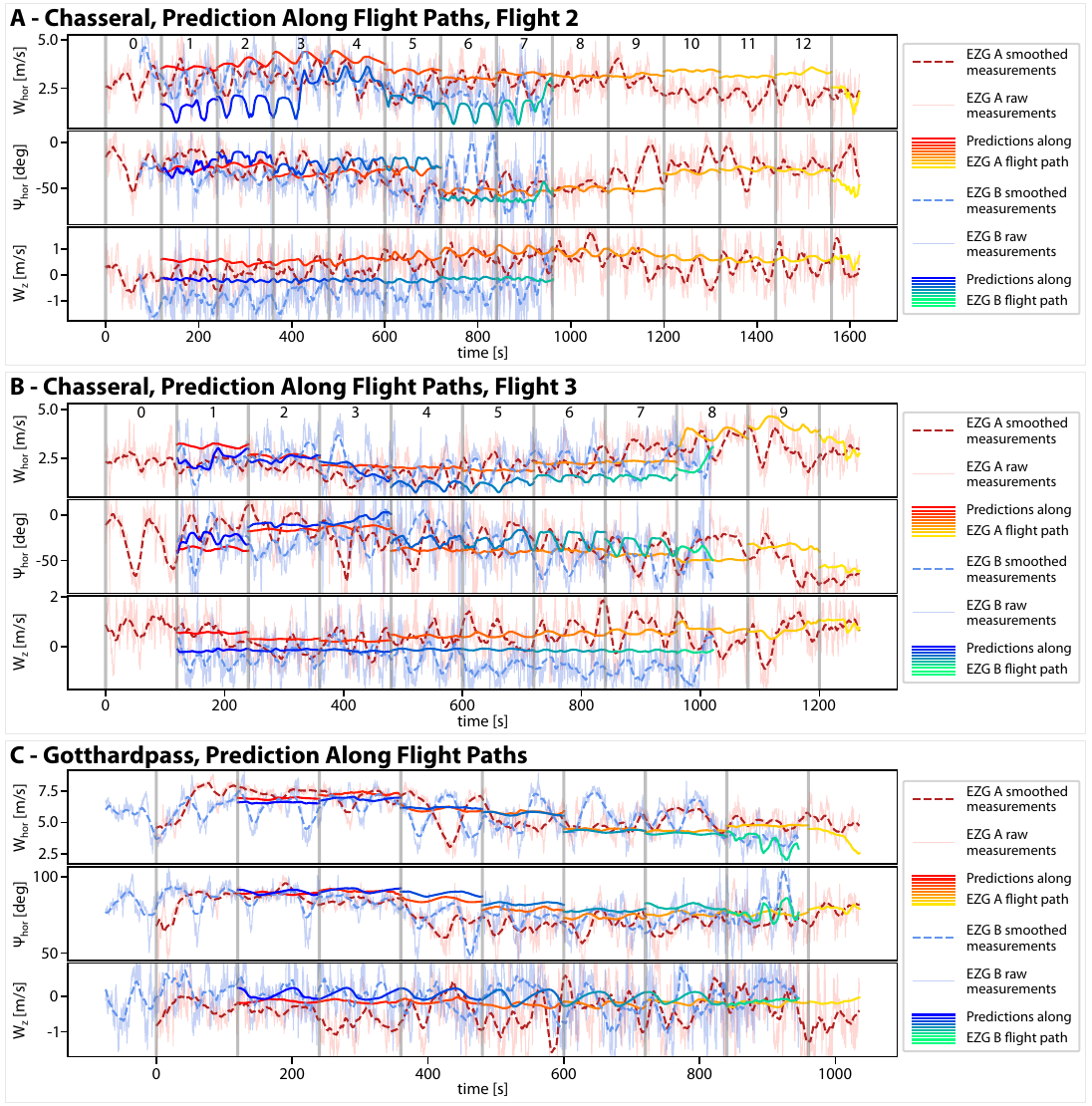}
    \caption[Airflow Calibration]{\textbf{\Ac{sUAV} flight results}. Additional sliding window prediction results for the second (\textbf{A}) and third (\textbf{B}) Chasseral flights and the Gotthardpass flight (\textbf{C}).}
    \label{fig:flight_tests_appendix}
\end{figure}

\end{document}